\documentclass[12pt,oneside,a4paper]{article}
\usepackage{amsfonts}

\usepackage{graphicx}
\usepackage{amsmath}
\usepackage{subfigure}
\usepackage{authblk}
\usepackage{setspace}
\usepackage{geometry}
\date{}

\title{Single Image-based Head Pose Estimation with Spherical Parametrization and 3D Morphing\footnote{\textbf{Funds}:This work was supported in part by the National Natural Science Foundation of China under Grants 61571274 and 61871342, in part by the Shandong Natural Science Funds for Distinguished Young Scholar under Grant JQ201614, in part by the Shandong Provincial Key Research and Development Plan under Grant 2017CXGC1504, and in part by the Young Scholars Program of Shandong University (YSPSDU) under Grant 2015WLJH39.}}

\author[a]{Hui Yuan \thanks{Corresponding author: huiyuan@sdu.edu.cn}}
\author[b]{Mengyu Li}
\author[c]{Junhui Hou}
\author[d]{Jimin Xiao}
\affil[a]{\scriptsize School of Control Science and Engineering, Shandong University, Ji'nan, China}
\affil[b]{\scriptsize School of Information Science and Engineering, Shandong University, Ji'nan, China}
\affil[c]{\scriptsize Department of Computer Science, City University of Hong Kong, Kowloon, Hong Kong}
\affil[d]{\scriptsize Department of Electrical and Electronic Engineering, Xi'an Jiaotong-Liverpool University, Suzhou, China}
\begin{document}
\maketitle

\doublespacing
\begin{abstract}
Head pose estimation plays a vital role in various applications, e.g., driver-assistance systems, human-computer interaction, virtual reality technology, and so on. We propose a novel geometry-based method for accurately estimating the head pose from a single 2D face image at a very low computational cost. Specifically, the rectangular coordinates of \emph{only four} non-coplanar feature points from a predefined 3D facial model as well as the corresponding ones automatically/manually extracted from a 2D face image are first normalized to exclude the effect of external factors (i.e., scale factor and translation parameters). Then, the four normalized 3D feature points are represented in spherical coordinates with reference to the uniquely determined sphere by themselves. Due to the spherical parametrization, the coordinates of feature points can then be morphed along all the three directions in the rectangular coordinates effectively. Finally, the rotation matrix indicating the head pose is obtained by minimizing the Euclidean distance between the normalized 2D feature points and the 2D re-projections of the morphed 3D feature points. Comprehensive experimental results over two popular datasets, i.e., \emph{Pointing'04} and \emph{Biwi Kinect}, demonstrate that the proposed method can estimate head poses with higher accuracy and lower run time than state-of-the-art geometry-based methods. Even compared with start-of-the-art learning-based methods or geometry-based methods with additional depth information, our method still produces comparable performance.
\end{abstract}




\section{Introduction}
Head pose estimation is strongly relevant with human-computer interaction, such as driver-assistance systems, human behavior analysis, virtual reality (VR)/ augment reality (AR)-based entertainment/education/telepresence. In a driver-assistance system, head pose of the driver is monitored so as to remind the driver to pay attention. For human behavior analysis, head pose estimation is used to assist the estimation of human gaze or face recognition, so as to accurately infer the intentions, desires, feelings, etc., of a person. For VR/AR applications, the desired field of view (FOV) of users can be estimated by head pose estimation. Head pose can be predicted by sensors embedded in head-mounted devices \cite{Ref1} or attached under the skin \cite{Ref2}, which are costly and annoying. Therefore, computer vision-based head pose estimation with high accuracy and in real-time is considered. Compared with sensor-based head pose estimation, it is more technologically challenging for computer vision-based head pose estimation, because the accuracy can be affected by many factors \cite{Ref3}, e.g., camera distortion, multisource non-Lambertian reflectance, facial expression, and the presence of accessories like glasses and hats, etc.

Computer vision-based head pose estimation needs to transform captured 2D face  images into a high level concept of directions \cite{Ref4}, i.e., three Euler angles: $\theta_{x}$(Pitch), $\theta_{y}$(Yaw) and $\theta_{z}$(Roll), as shown in Fig. 1. According to \cite{Ref5}, we can classify existing computer vision-based head pose estimation methods into two categories: learning-based methods \cite{Ref6}-\cite{Ref20} and geometry-based methods \cite{Ref21}-\cite{Ref29}, see section II for details.

\begin{figure}
    \centering
    \includegraphics[width=4cm]{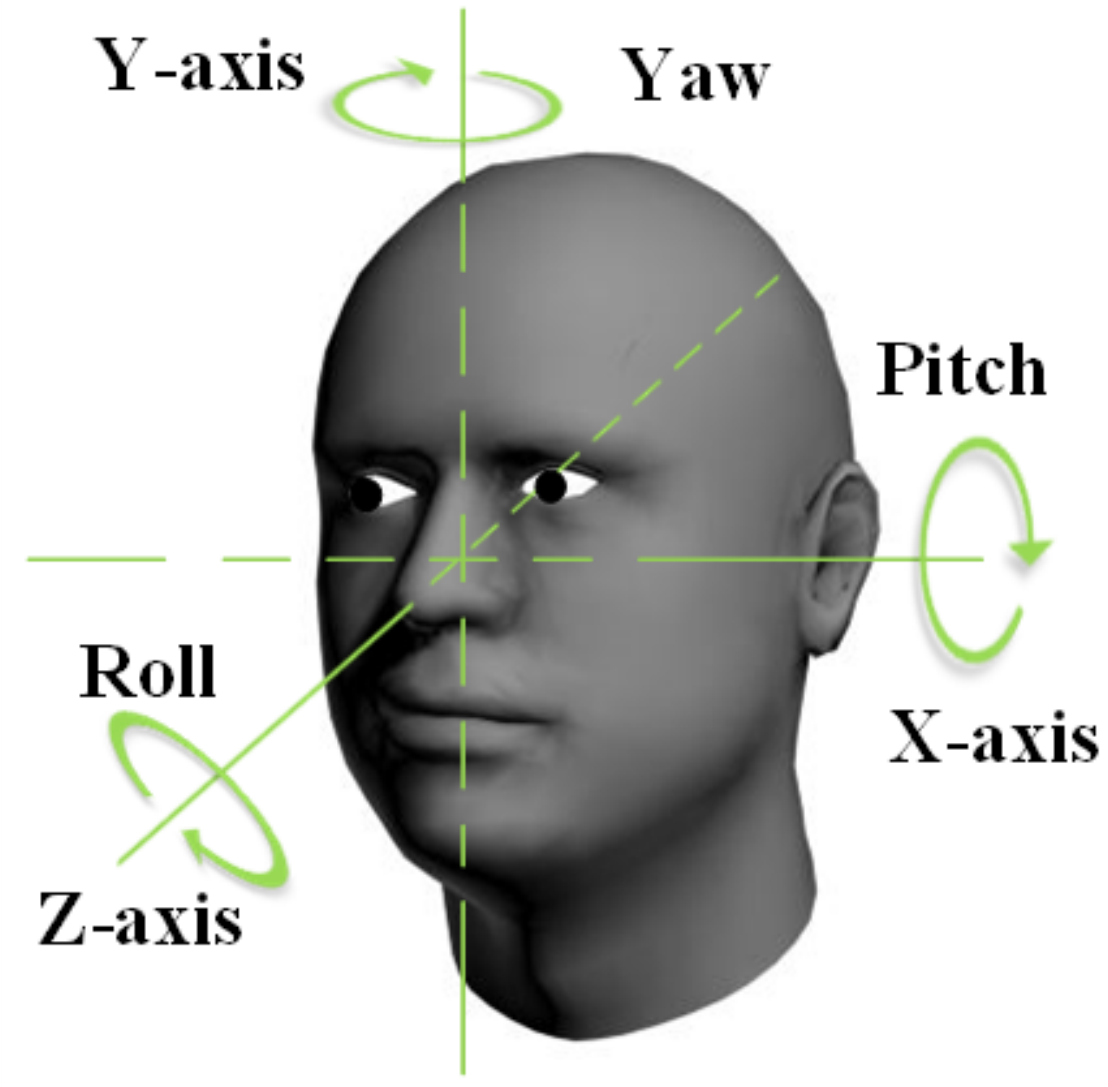}
    \caption{Head pose represented by three angles, i.e., $\theta_{x}$(Pitch), $\theta_{y}$(Yaw) and $\theta_{z}$(Roll).}
    \label{fig1}
\end{figure}

\begin{figure}[htbp]
    \centering
    \includegraphics[width=12cm]{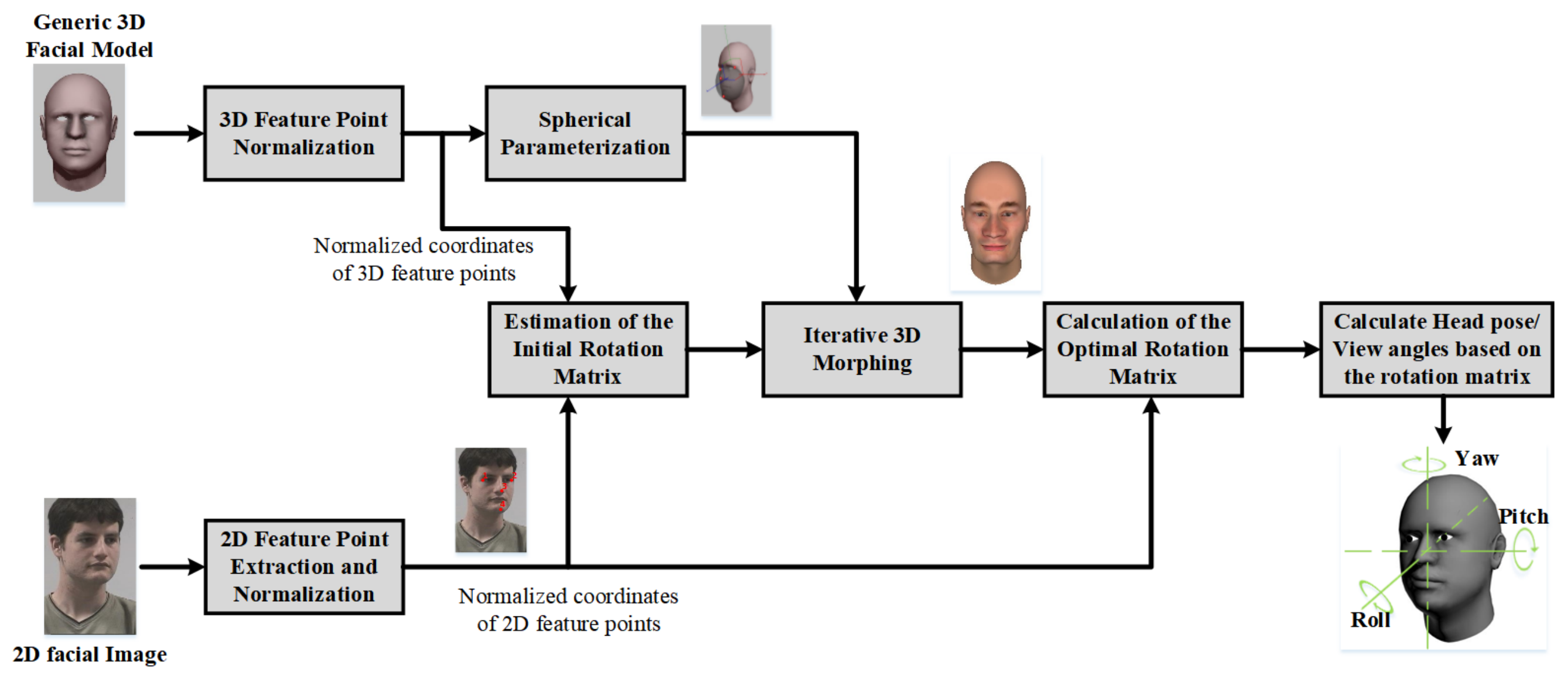}
    \caption{Illustration of flowchart of the proposed method.}
    \label{fig2}
\end{figure}

In this paper, as shown in Fig. 2, we propose an accurate geometry-based method for estimating the head pose from a \emph{single 2D face image} with a low computational cost. Specifically, the rectangular coordinates of \emph{\textbf{only four}} non-coplanar feature points from a predefined 3D facial model as well as the corresponding ones automatically/manually extracted from a 2D face image are firstly normalized to eliminate the effect of scale factor and translation parameters. Accordingly, the geometric relationship between the normalized coordinates of feature points in 2D face image and those in 3D facial model can be built by only using a rotation matrix that is denoted by $\textbf{\emph{R}}\in\mathbb{R}^{3\times3}$. In order to adapt the various individuals, the four normalized 3D feature points are then represented in spherical coordinates with reference to the uniquely determined sphere by themselves, which are further iteratively refined by varying the azimuth and elevation in the spherical coordinates. \textbf{In this way, the rectangular coordinates of the 3D facial model could be morphed along all the three directions to better adapt different faces of individuals.} Finally, the rotation matrix indicating the head pose is obtained by minimizing the Euclidean distance between the normalized 2D feature points and the 2D re-projections of morphed 3D feature points. The contributions and novelties of this paper are summarized as follows.

(1) We propose a geometry-based method, which is capable of estimating the 3D head pose from a single 2D face image accurately and at a very low computational cost.

(2) Our method only needs four non-coplanar 2D feature points, which can reduce the effect of inaccurate 2D feature points and the computational complexity.

(3) We propose an efficient 3D morphing method with spherical parameterization, which is able to deform a pre-defined 3D facial model along all the three directions of the rectangular coordinates.

Experimental results over two popular datasets, i.e., \emph{Pointing'04} and \emph{Biwi Kinect}, demonstrate that the proposed method can estimate head poses with higher accuracy and lower run time than the state-of-the-art methods.

The rest of this paper is organized as follows. In Section II, related works are given. In Section III, the preliminary of the 3D facial model-based head pose estimation is presented. Details of the proposed method are given in Section IV. Experimental results and conclusions are given in Section V and VI respectively.

\section{Related Works}
\subsection{Learning-based methods}
Learning-based methods attempt to find the matching relationship between query face images (usually represented by extracted appearance features) and head positions with the support of huge face datasets in which a sufficient amount of training data uniformly distributed across various pose angles is provided.

Mathematically, head poses are estimated by solving a regression or classification problem in the learning-based methods. Chutorian \emph{et al}. \cite{Ref6} trained two support vector regression models based on Localized Gradient Orientation histograms to match the orientation of the driver's head. Fu \emph{et al}. \cite{Ref7} categorized the head pose into 12 gaze zones based on facial features and then used self-learning algorithm and particle filter to estimate the head poses.  Ba and Odobez\cite{Ref8} applied a Bayesian probabilistic framework that is solved through particle filtering techniques for head tracking and pose estimation. Tan \emph{et al}. \cite{Ref9} presented a random forest-based framework to estimation head pose from depth images. Liang \emph{et al}. \cite{Ref10} proposed an improved Hough-voting with random forest via simultaneously changing the weights of leaf votes with L0-regularized optimization and pruning unreliable leaf nodes in the forest. Drouard \emph{et al}. \cite{Ref11} used a mixture of linear regression with partially-latent output that learns to map high-dimensional feature vectors (extracted from bounding boxes of faces) onto the joint space of head-pose angles and bounding-box shifts by unsupervised manifold learning so that they can be robustly predicted in the presence of unobservable phenomena. Rajagopal \emph{et al}. \cite{Ref12} explored transfer learning approaches for efficient multi-view head pose classification with minimal target training data. Asteriadis \emph{et al}. \cite{Ref13} combined two novel techniques of distance vector fields (DVFs) and convolutional neural network (CNN), respectively based on a local information and holistic appearance information to estimate head pose angles. Sadeghzadeh \emph{et al}. \cite{Ref14} trained a fuzzy system with the input of the facial geometry features, such as the ratios and angles among these feature points, to estimate head pose angles. Riegler \emph{et al}. \cite{Ref15} proposed a Hough Networks (HNs) which combines the Hough Forests with CNN by performing classification and regression simultaneously. Papazov \emph{et al}. \cite{Ref16} introduced a surface patch descriptor-based method for head pose estimation using depth and color information. Recently, N. Ruiz, E. Chong, and J. M. Rehg \cite{Ref17}, presented an elegant and robust way to determine head pose by training a multi-loss CNN (namely HopeNet) to predict the yaw, pitch and roll angles directly from image intensities through joint binned pose classification and regression. Tsun-Yi Yang, \emph{et al.} \cite{Ref18} proposed a soft stagewsie regression network (SSR-Net) for age estimation, which can also be used for head pose estimation directly because of its efficient facial feature detection. Based on the SSR-Net structure, T.-Y. Yang \emph{et al.} also proposed to learn a fine-grained structure mapping (FSA-Net) for spatially grouping features before aggregation to predict head pose efficiently in \cite{Ref19}. F. Kuhnke and J. Ostermann \cite{Ref20} proposed an efficient head pose estimation method by using synthetic images and a partial domain adversarial networks.

\subsection{Geometry-based methods}
Geometry-based methods estimate the head pose by geometrical calculation with feature points. They can be further divided into 2 categories coarsely, i.e. geometry distribution-based methods \cite{Ref21}-\cite{Ref25} and 3D facial model-based methods\cite{Ref26}-\cite{Ref30}.

Geometry distribution-based methods attempt to estimate the head poses directly from the geometry distribution of feature points on a 2D face image based on a fixed geometrical model. At the early stage, Gee \emph{et al}. \cite{Ref21} compared the proportion between five facial feature points and the length of nose with a fixed value to determine the head direction. Narayanan \emph{et al}. \cite{Ref22} adopted an ellipse-circle model to estimate the head pose. Nikolaidis \emph{et al}. \cite{Ref23} distorted the isosceles triangle formed by the two eyes and the mouth to estimate the head yaw angle. In order to estimate more robust and accurate yaw angle, Narayanan \emph{et al}. \cite{Ref24} proposed an improved generic geometric model-based on cylindrical and ellipsoidal models to customize the head pose into 12 different models. However, the distribution of feature points of human faces varies a lot because of gender, races, ages, etc. Therefore, it is hard to estimate various head poses accurately by a fixed geometrical model.

3D facial model-based methods estimate the head poses from the correspondence between feature points on a 2D face image and those on a 3D facial model which can be morphed to match individual facial image. By looking for the projection relation between a 3D facial model and a 2D face image, head pose angles can be calculated from the elements in the rotation matrix directly (see Section III for details).  Fridman \emph{et al}. \cite{Ref25} solved the rotation matrix to estimate the head pose according to a 3D facial model and the corresponding 2D facial feature points directly. Martins \emph{et al}. \cite{Ref26} proposed a real-time 3D facial model updating method for head pose estimation in which the iterative closest point (ICP) algorithm is also used to find the best matching pair of a 2D face image and a 3D facial model. Li \emph{et al}. \cite{Ref27} proposed a real-time face template reconstruction algorithm based on head pose estimation and tracking in which the ICP algorithm is also used. Meyer \emph{et al}. \cite{Ref28} combined particle swarm optimization (PSO) and the ICP algorithm to estimate the head pose. However, all of them \cite{Ref25}-\cite{Ref28} have used the depth information of 2D face images. Kong  and Mbouna \cite{Ref29} estimated head pose angles from a single 2D face image using a 3D facial model morphed from a reference facial model. But the 3D facial model is morphed only along with one direction.

\section{Preliminary: Head Pose Representation and 3-D Projection}
As shown in Fig. 1, a head can typically rotate around X, Y, and Z axes, and thus, head pose can be represented by three Euler angles, i.e., $\theta_{x}$(Pitch), $\theta_{y}$(Yaw) and $\theta_{z}$(Roll). The three angles correspond to the head actions of nodding, shaking, and rolling, which are helpful for human behavior analysis, gaze and FOV estimation, etc. In a 3D world coordinate, when a point located at (\emph{x}, \emph{y}, \emph{z}) rotates $\theta_{x}$ around the X axis, the resulted coordinate of the point will be
\begin{equation}\label{E1}
    (\emph{\large x}_{\emph{\scriptsize X}}\ \emph{\large y}_{\emph{\scriptsize Y}}\
    \emph{\large z}_{\emph{\scriptsize Z}})=\textbf{\emph{R}}_{X}\cdot(\emph{x}\
    \emph{y}\ \emph{z})^\emph{\scriptsize T},
\end{equation}
where
\begin{equation}\label{E2}
\textbf{\emph{R}}_{X} = \left[
\begin{array}{ccc}
1&0&0\\
0&{\cos {\theta _x}}&{ - \sin {\theta _x}}\\
0&{\sin {\theta _x}}&{\cos {\theta _x}}
\end{array}
\right].
\end{equation}
Similarly, when the point rotates $\theta_{y}$ and $\theta_{z}$ around the Y and Z axes, the resulted coordinate will be
\begin{equation}\label{E3}
    ({{\emph{\large x}_\emph{{\scriptsize Y}}}}\ {{\emph{\large y}_\emph{{\scriptsize Y}}}}\ {{\emph{\large z}_\emph{{\scriptsize Y}}}})^{\emph{\scriptsize T}}
    = \textbf{\emph{R}}_{Y} \cdot (x\ y\ z)^{\emph{\scriptsize T}},
\end{equation}
and
\begin{equation}\label{E4}
    ({{\emph{\large x}_\emph{{\scriptsize Z}}}}\ {{\emph{\large y}_\emph{{\scriptsize Z}}}}\ {{\emph{\large z}_\emph{{\scriptsize Z}}}})^{\emph{\scriptsize T}}
    = \textbf{\emph{R}}_{Z} \cdot (x\ y\ z)^{\emph{\scriptsize T}},
\end{equation}
respectively, where
\begin{equation}\label{E5}
\textbf{\emph{R}}_{Y} = \left[
\begin{array}{ccc}
{\cos {\theta _y}}&0&{\sin {\theta _y}}\\
0&1&0\\
-{\sin {\theta _y}}&0&{\cos {\theta _y}}
\end{array}
\right],
\end{equation}
and
\begin{equation}\label{E6}
\textbf{\emph{R}}_{Z} = \left[
\begin{array}{ccc}
{\cos {\theta _z}}&-{\sin {\theta _z}}&0\\
{\sin {\theta _z}}&{\cos {\theta _z}}&0\\
0&0&1
\end{array}
\right].
\end{equation}
Therefore, for any kind of rotation, the resulted coordinate of the point can be written as:
\begin{equation}\label{E7}
    ({{\emph{\large x}_\emph{{\scriptsize X\scriptsize Y\scriptsize Z}}}}\ {{\emph{\large y}_\emph{{\scriptsize X\scriptsize Y\scriptsize Z}}}}\ {{\emph{\large z}_\emph{{\scriptsize X\scriptsize Y\scriptsize Z}}}})^{\emph{\scriptsize T}}
    = \textbf{\emph{R}}_{X}\textbf{\emph{R}}_{Y}\textbf{\emph{R}}_{Z} \cdot (x\ y\ z)^{\emph{\scriptsize T}}=\emph{\textbf{R}} \cdot (x\ y\ z)^{\emph{\scriptsize
    T}}.
\end{equation}

\newcounter{mytempeqncnt}
\begin{figure*}[hb]
\normalsize \setcounter{mytempeqncnt}{\value{equation}}
\setcounter{equation}{7}
\begin{equation}\label{E8}
\begin{split}
\textbf{\emph{R}} &= \left[
\begin{array}{ccc}
{r_{11}}&{r_{12}}&{r_{13}}\\
{r_{21}}&{r_{22}}&{r_{23}}\\
{r_{31}}&{r_{32}}&{r_{33}}
\end{array}
\right]
\\&= \left[
\begin{array}{ccc}
{\cos {\theta _z}\cos {\theta _y}}&{\cos {\theta
_z}\sin{\theta_y}\sin {\theta _x}-\sin {\theta _z}\cos {\theta
_x}}&{\cos {\theta _z}\sin {\theta _y}\cos {\theta _x}+\sin {\theta
_z}\sin {\theta _x}}\\
{\sin {\theta _z}\cos {\theta _y}}&{\sin {\theta
_z}\sin{\theta_y}\sin {\theta _x}+\cos {\theta _z}\cos {\theta
_x}}&{\sin {\theta _z}\sin {\theta _y}\cos {\theta _x}-\cos {\theta
_z}\sin {\theta _x}}\\
{-\sin {\theta _y}}&{\cos {\theta _y}\sin {\theta _x}}&{\cos {\theta
_y}\cos {\theta _x}}
\end{array}
\right].
\end{split}
\end{equation}
\setcounter{equation}{\value{mytempeqncnt}} \hrulefill \vspace*{4pt}
\end{figure*}

For the rotation matrix \textbf{\emph{R}}, as shown in Eq. (8), the three row vectors $\textbf{\emph{r}}_\textbf{1}=({r_{11}}\ {r_{12}}\
{r_{13}})^T,\textbf{\emph{r}}_\textbf{2}=({r_{21}}\ {r_{22}}\
{r_{23}})^T and\ \textbf{\emph{r}}_\textbf{3}=({r_{31}}\ {r_{32}}\
{r_{33}})^T$ have the following relations:
\begin{equation}\label{E9}
\setcounter{equation}{9} \left\{ \begin{array}{l}
\textbf{\emph{r}}_\textbf{1}^T\textbf{\emph{r}}_\textbf{2} = 0\\
\textbf{\emph{r}}_\textbf{2}^T\textbf{\emph{r}}_\textbf{1} = 0\\
\textbf{\emph{r}}_\textbf{1}^T\textbf{\emph{r}}_\textbf{1} = 1\\
\textbf{\emph{r}}_\textbf{2}^T\textbf{\emph{r}}_\textbf{2} = 1\\
\textbf{\emph{r}}_\textbf{3} =
\textbf{\emph{r}}_\textbf{1}\times\textbf{\emph{r}}_\textbf{2}{.}
\end{array} \right.
\end{equation}
Accordingly, the Euler angles $\theta_{x}, \theta_{y}$, and $\theta_{z}$ can be calculated as:
\begin{equation}\label{E10}
\setcounter{equation}{10} \left\{ \begin{array}{l}
\theta_{x}={\tan^{-1}}\frac{{{r_{32}}}}{{{r_{33}}}}\\
\theta_{y}=-{\tan ^{-1}}\frac{{{r_{31}}}}{{\sqrt {r_{32}^2 + r_{33}^2} }}\\
\theta_{z}={\tan ^{-1}}\frac{{{r_{21}}}}{{{r_{11}}}}{.}
\end{array} \right.
\end{equation}

According to the pinhole camera model, the projection from a 3D facial model to a 2D face image plane can be described as:
\begin{equation}\label{E11}
\setcounter{equation}{11} s\!\left(\! {\begin{array}{c}
\mu\\
\nu\\
1
\end{array}}\! \right)=\!\textbf{\emph{A}}[\textbf{\emph{R}}\
\textbf{\emph{t}}]=\!\left[\!
\begin{array}{ccc}
\alpha&\gamma&\mu_0\\
0&\beta&\nu_0\\
0&0&1
\end{array}\!
\right]\!\left[\! {\begin{array}{c}
\textbf{\emph{r}}_\textbf{1}^Tt_1\\
\textbf{\emph{r}}_\textbf{2}^Tt_2\\
\textbf{\emph{r}}_\textbf{3}^Tt_3
\end{array}}\! \right]
\!\left(\! {\begin{array}{c}
x\\
y\\
z\\
1
\end{array}} \!\right) {,}
\end{equation}
where \emph{s} is a scale factor commonly denoted as the projective depth, $(\mu,\nu)$ is the projected pixel coordinate in the image plane, \textbf{\emph{A}} is the intrinsic parameter of camera, $\alpha$ and $\beta$ are scaling factors of the two image axes that depend on the horizontal and vertical focal length of the camera lens, $\gamma$ is the skew, $(\mu_0,\nu_0)$ is the coordinate of principal point, and $\textbf{\emph{t}}=(\emph{t}_1\ \emph{t}_2\ \emph{t}_3)^T$ is the translation vector denoting the distance between camera center and object center with respect to the 3D world coordinate system.

In an ideal camera system without any distortion, we have $\gamma=0, \alpha=\beta, and\ \mu_0=\nu_0=0$. Furthermore, based on the affine perspective assumption \cite{Ref30}, when the distance of an object from a camera (corresponding to \emph{s} and $t_3$) is much larger than the depth variation (corresponding to the elements in the rotation matrix \textbf{\emph{R}}) of the object, Eq. (11) can be simplified as:
\begin{equation}\label{E12}
\setcounter{equation}{12} \left(\! {\begin{array}{c}
\mu\\
\nu\\
1
\end{array}}\! \right)=\!\left[\! {\begin{array}{cccc}
\frac{\alpha }{s}{r_{11}}&\frac{\alpha }{s}{r_{12}}&\frac{\alpha }{s}{r_{13}}&\frac{\alpha }{s}{t_{1}}\\
\frac{\alpha }{s}{r_{21}}&\frac{\alpha }{s}{r_{22}}&\frac{\alpha }{s}{r_{23}}&\frac{\alpha }{s}{t_{2}}\\
0&0&0&1
\end{array}}\! \right]
\!\left(\! {\begin{array}{c}
x\\
y\\
z\\
1
\end{array}} \!\right).
\end{equation}
Subsequently, we have
\begin{equation}\label{E13}
\begin{split}
\setcounter{equation}{13} \!\left(\! {\begin{array}{c}
\mu\!-\!s{'}t_{1}\\
\nu\!-\!s{'}t_{2}
\end{array}}\!\right)\!&=\!s{'}\!\left[\!
\begin{array}{ccc}
{r_{11}}\ {r_{12}}\ {r_{13}}\\
{r_{21}}\ {r_{22}}\ {r_{23}}\\
\end{array}
\!\right]\!\left(\! {\begin{array}{c}
x\\
y\\
z
\end{array}} \!\right)\! =\!s{'}\!\left[\! \begin{array}{l}
\textbf{\emph{r}}_\textbf{1}^T\\
\textbf{\emph{r}}_\textbf{2}^T
\end{array} \!\right]\!\left(\! {\begin{array}{c}
x\\
y\\
z
\end{array}} \!\right),
\end{split}
\end{equation}
where $s{'}={\raise0.7ex\hbox{$\alpha $} \!\mathord{\left/
 {\vphantom {\alpha  s}}\right.\kern-\nulldelimiterspace}
\!\lower0.7ex\hbox{$s$}}$.

\section{Proposed Method}
According to Eq. (13), with 5 matching pairs of feature points from a 2D face image and a predefined 3D facial model, one can intuitively obtain all the elements of the rotation matrix by solving linear equations. However, such an intuitive way may produce inaccurate rotation matrix because the pre-defined 3D facial model cannot exactly adapt to all individuals with various genders, races, ages, and facial expressions. To improve the accuracy of the estimated rotation matrix, one alternative method is to take more matching pairs of feature points and adopts more advanced regression methods. However, the complexity of extracting a large amount of high-quality matching pairs is very high. Moreover, the head pose angles are only determined by the rotation matrix, and thus the effect of the external factors in Eq. (13) (i.e., the scale factor and translation parameters) has to be excluded. Based on these analyses, we propose a head pose estimation method only based on geometry, in which \emph{only four} non-coplanar feature points are employed. In the proposed method, the effect introduced by the external factors is removed by coordinate normalization that is similar to \cite{Ref29}; afterwards, the 3D morphing method with spherical parametrization is proposed to morph the coordinates of the 3D facial model along all the three directions in the rectangular coordinates to adapt the 3D facial model to each individual.

\subsection{Feature Point Normalization}

Let ${\textbf{\emph{m}}_i} = ({\mu _i}\ {\nu _i})^T$ and $\textbf{\emph{M}}_i=(x_{i}\ y_{i}\ z_{i})^T$ denote the coordinates of the \emph{i}-th (\ i\ $\in\{1,2,\cdots,N\}$\ ) 2D and 3D feature points respectively, which meet Eq. (13). We further rewrite Eq.(13) in its matrix form:
\begin{equation}\label{E14}
\emph{\textbf{m}}_{\emph{i}}-\emph{s}'\emph{\textbf{t}}=\emph{s}'\emph{\textbf{R}}'\emph{\textbf{M}}_{\emph{i}},
\end{equation}
where $\textbf{\emph{R}}'=[\textbf{\emph{r}}_1\
\textbf{\emph{r}}_2]^T$. Then, we normalize the coordinates of 2D
and 3D feature points separately:
\begin{equation}\label{E15}
\setcounter{equation}{15}
\textbf{\emph{m}}{'}_{i}=\frac{\textbf{\emph{m}}_i-\textbf{\emph{m}}_0}{{\left\|
{\textbf{\emph{m}}_i-\textbf{\emph{m}}_0} \right\|_2}}
\end{equation}
\begin{equation}\label{E16}
\setcounter{equation}{16}
\textbf{\emph{M}}{'}_{i}=\frac{\textbf{\emph{M}}_i-\textbf{\emph{M}}_0}{{\left\|
{\textbf{\emph{M}}_i-\textbf{\emph{M}}_0} \right\|_2}}
\end{equation}
where ${{\textbf{\emph{m}}{'}}_i}=({{\mu{'}} _i}\ {{\nu{'}} _i})^T$ and ${\textbf{\emph{M}}{'}}_i=(x{'}_{i}\ y{'}_{i}\ z{'}_{i})^T$ stand for the normalized $\textbf{\emph{m}}_i$ and $\textbf{\emph{M}}_i$, respectively, $\textbf{\emph{m}}_0$ and $\textbf{\emph{M}}_0$ are the centroids of \emph{N} 2D and 3D feature points:
\begin{equation}\label{E17}
\setcounter{equation}{17} {\textbf{\emph{m}}_0} =
\frac{1}{N}\sum\nolimits_{i = 1}^N {{\textbf{\emph{m}}_i}},
\end{equation}
\begin{equation}\label{E18}
\setcounter{equation}{18} {\textbf{\emph{M}}_0} =
\frac{1}{N}\sum\nolimits_{i = 1}^N {{\textbf{\emph{M}}_i}}.
\end{equation}
By substituting Eqs. (15) and (16) into Eq. (14), we have
\begin{equation}\label{E19}
\setcounter{equation}{19} {\textbf{\emph{m}}}
{'}_i=\textbf{\emph{R}}{'}{\textbf{\emph{M}}}{'}_i{,}
\end{equation}
From Eq. (19), we can observe that the translation parameter $\textbf{\emph{t}}=(\emph{t}_{1},\emph{t}_{2} )^{T}$ and scale factor $\emph{s}'$ are eliminated making it possible to estimate the rotation matrix from at least 4 matching pairs of 2D and 3D feature points. The detailed derivation of Eq. (19) from Eq. (14) with normalized coordinates is theoretically proven in \emph{Appendix I}.

\subsection{Rotation Matrix Estimation using 3D Morphing with Spherical Parametrization}
With the normalized 2D and 3D feature points, the rotation matrix can be obtained directly by solving Eq. (19), i.e., $\textbf{\emph{R}}' = {\textbf{\emph{m}}'_i}{\textbf{\emph{M}}'_i}^T{({{{\textbf{\emph{M}}'}_i}{\textbf{\emph{M}}'_i}^T})}^{-1}$. However, as aforementioned, inaccurate estimation will be resulted in by using only one pre-defined 3D facial model to fit individuals with various genders, races, ages, and facial expressions. To this end, we propose to iteratively refine the \emph{four} normalized non-coplanar 3D feature points using 3D morphing with spherical parameterization, and the refined 3D feature points can adapt to each individual better for more accurate estimation. Owing to the advantage of the spherical parameterization of the 3D facial model, the proposed method can morph all the three directions without introducing too much computational complexity to adapt each individual flexibly and accurately.

 The four normalized 3D feature points can uniquely determine a sphere (see \emph{Appendix} 2). Let  $\textbf{\emph{B}}'_i=(l\ \phi_i\ \varphi_i)^T$ denote the corresponding spherical coordinate of the \emph{i}-th 3D feature point, where \emph{l} is the radius of the sphere, $\phi_i$\ and\ $\varphi_i$ are the azimuth and elevation, respectively. The relationship between $\textbf{\emph{B}}'_i=(l\ \phi_i\ \varphi_i)^T$ and ${\textbf{\emph{M}}{'}}_i=(x{'}_{i}\ y{'}_{i}\ z{'}_{i})^T$ is expressed as
\begin{equation}\label{E20}
\setcounter{equation}{20} \left\{ \begin{array}{l}
{{x'}_i} - {{x'}_o} = l\sin {\varphi _i}\cos {\phi _i}\\
{{y'}_i} - {{y'}_o} = l\sin {\varphi _i}\sin {\phi _i}\\
{{z'}_i} - {{z'}_o} = l\cos {\varphi _i}{,}
\end{array} \right.
\end{equation}
where $(x{'}_{o}\ y{'}_{o}\ z{'}_{o})^T$ is the rectangular coordinate of the center of the sphere. Therefore, individual faces with different characteristics can be adapted by changing the position of these 3D feature points on the sphere, i.e., morphing azimuth ($\phi_i$) and elevation ($\varphi_i$). For the \emph{i}-th 3D feature point with the morphing parameters denoted as, ${\mathbf{\Delta}_i} = {(0\ {{\Delta_{i,\phi}}}\ {{\Delta_{i,\varphi}}})^T}$, the morphed spherical coordinate is represented as ${\hat {\textbf{\emph{B}}}'_i} = {\textbf{\emph{B}}'_i} + {\mathbf{\Delta}_i}$, and the corresponding rectangular coordinate ${\hat {\textbf{\emph{M}}}'_i} =(\hat x'_i\ \hat y'_i\ \hat z'_i)^T$ can be obtained:
\begin{equation}\label{E21}
\setcounter{equation}{21} \left\{ \begin{array}{l}
{\hat x'_i} - {{x'}_o} = l\sin {(\varphi _i}+{\Delta _{i,\varphi }})\cos {(\phi _i+{\Delta _{i,\phi }})}\\
{\hat y'_i} - {{y'}_o} = l\sin {(\varphi _i+{\Delta _{i,\varphi }})}\sin {(\phi _i+{\Delta _{i,\phi }})}\\
{\hat z'_i} - {{z'}_o} = l\cos {(\varphi _i+{\Delta _{i,\varphi
}})}{.}
\end{array} \right.
\end{equation}

Let $\hat{\textbf{\emph{M}}}'_{i,1}$ denote the initial rectangular coordinate derived from the initial morphing parameter ${\mathbf{\Delta}_{i,1}}$, and the initial re-projected 2D coordinates of $\hat{\mathbf{\emph{\textbf{M}}}}'_{i,1}$ be $\hat{\textbf{\emph{m}}}'_{i}(\textbf{\emph{R}}^{1},\hat{\textbf{\emph{M}}}'_{i,1})$ in which the initial rotation matrix
$\textbf{\emph{R}}^{1}=[\textbf{\emph{r}}^1_1\
\textbf{\emph{r}}^1_2\ \textbf{\emph{r}}^1_3]^T$ can be calculated by solving

\begin{equation}\label{E22}
\setcounter{equation}{22} \left\{ \begin{array}{l}
\textbf{\emph{m}}'_i=[\textbf{\emph{r}}^1_1\
\textbf{\emph{r}}^1_2]^T\textbf{\emph{M}}'_i\\
\textbf{\emph{r}}^1_3=\textbf{\emph{r}}^1_1\times\textbf{\emph{r}}^1_2.
\end{array} \right.
\end{equation}
In order to adapt individual faces by morphing the 3D facial model, the discrepancy between $\hat{\textbf{\emph{m}}}'_{i}(\textbf{\emph{R}}^{1},\hat{\textbf{\emph{M}}}'_{i,1})$ and $\textbf{\emph{m}}'_i$ can be calculated and then minimized iteratively to find the best morphing parameters ${\mathbf{\Delta}
_{i,opt}}$:
\begin{equation}\label{E23}
\setcounter{equation}{23} \underbrace {\min }_{{\mathbf{\Delta}
_i}}\sum\nolimits_{i = 1}^N {\left\| {{{\textbf{\emph{m}}'}_i} -
{{\hat {\textbf{\emph{m}}}'}_i}({\textbf{\emph{R}}^1},{{\hat
{\textbf{\emph{M}}}'}_{i,k}})} \right\|} _2^2,
\end{equation}
where $k\in\{1,\ldots,opt\}$ is the iteration index. To ensure that the morphed 3D model is still a human face, we penalize the Euclidean distance between the initial 3D feature points and the morphed ones so that the morphing parameters cannot be arbitrary values, leading to

\begin{equation}\label{E24}
\begin{split}
\setcounter{equation}{24} \underbrace {\min }_{{\mathbf{\Delta}
_i}}\sum\nolimits_{i = 1}^N &{\left\| {{{\textbf{\emph{m}}'}_i} -
{{\hat {\textbf{\emph{m}}}'}_i}({\textbf{\emph{R}}^1},{{\hat
{\textbf{\emph{M}}}'}_{i,k}})} \right\|} _2^2 + \eta \sum\nolimits_{i = 1}^N {\left\| {{{\hat
{\textbf{\emph{M}}}'}_{i,k}} - {{\hat {\textbf{\emph{M}}}'}_{i,1}}}
\right\|} _2^2{,}
\end{split}
\end{equation}
where $\eta$ is a penalty parameter that controls the deformation of the 3D facial model. The non-linear least squares problem in Eq. (24) can be converted to a linear least squares problem by removing the high order terms of its Taylor expansion with respect to $\mathbf{\Delta}_i$. Then the Levenberg-Marquardt (LM) algorithm that requires a partial derivative of $\mathbf{\Delta}_i$ is employed to solve the minimization problem iteratively, in which each iteration aims to seek for a suitable damping factor based on the trust region method so as to acquire a reliable update value of $\mathbf{\Delta}_i$ and decrease the value of objective function to convergence. With the optimal morphing parameter $\mathbf{\Delta}_{i,opt}$ obtained, the optimal coordinates of 3D feature points in the rectangular coordinates, i.e., ${\hat {\textbf{\emph{M}}}'_{i,opt}} =(\hat x'_{i,opt}\ \hat y'_{i,opt}\ \hat z'_{i,opt})^T$, can be calculated by
\begin{equation}\label{E25}
\setcounter{equation}{25} \left\{\!\begin{array}{l}
{{\hat{x}'}_{i,opt}} - {{x'}_o} = l\sin {(\varphi _i}+{\Delta _{i,\varphi,opt }})\cos {(\phi _i+{\Delta _{i,\phi,opt }})}\\
{{\hat{y}'}_{i,opt}} - {{y'}_o} = l\sin {(\varphi _i+{\Delta _{i,\varphi,opt }})}\sin {(\phi _i+{\Delta _{i,\phi,opt }})}\\
{{\hat{z}'}_{i,opt}} - {{z'}_o} = l\cos {(\varphi _i+{\Delta
_{i,\varphi,opt }})}{.}
\end{array} \right.
\end{equation}

Finally, the optimal rotation matrix ${\textbf{\emph{R}}^{opt}} = {[
{\textbf{\emph{r}}_1^{opt}}\ {\textbf{\emph{r}}_2^{opt}}\
{\textbf{\emph{r}}_3^{opt}} ]^T}$ can be calculated by solving
\begin{equation}\label{E26}
\setcounter{equation}{26} \left\{ \begin{array}{l}
{{\hat {\textbf{\emph{M}}}'}_{i,opt}} = {{\textbf{\emph{M}}'}_i} + {\mathbf{\Delta}_{i,opt}}\\
{{\textbf{\emph{m}}'}_i} = {[ {\textbf{\emph{r}}_1^{opt}}\
{\textbf{\emph{r}}_2^{opt}}]^T}{{\hat {\textbf{\emph{M}}}'}_{i,opt}}\\
\textbf{\emph{r}}_3^{opt} = \textbf{\emph{r}}_1^{opt} \times
\textbf{\emph{r}}_2^{opt}{.}
\end{array} \right.
\end{equation}
 Then, the head pose angles can then be calculated by Eq. (10).

When solving Eq.(24), the termination condition is set as $\emph{E}(\emph{k}+1)-\emph{E}(\emph{k})\leq10^{-6}$, where \emph{E}(\emph{k}) represents the value of the objective function (24) at the \emph{k}-th iteration.

\section{Experimental Results}
To verify the performance of the proposed method, extensive experiments were conducted with a PC equipped with Intel Core i7-7700HQ CPU@2.8GHz, 8GB RAM, and Windows 10 64bits operating system. Two head pose datasets were used, i.e., \emph{Pointing'04} \cite{Ref31} and \emph{Biwi Kinect} \cite{Ref32}. The \emph{Pointing'04} dataset consists of 2790 images of 15 sets of human face with various races, genders, and ages, i.e., 12 sets of male Caucasians, 1 set of female Caucasian, 1 set of female Asian, and 1 set of male Indian. Each set contains of 2 series of 93 images with the size of 384$\times$288 at different poses angles. The yaw and pitch angles vary from -90 degree to +90 degree, and there is no roll angle in the \emph{Pointing'04} dataset. For the \emph{Biwi Kinect} dataset, there are 15678 images of size 640$\times$480 generated from 20 persons (14 males and 6 females). The head pose angles vary from -75 degree to +75 (yaw), -60 degree to +60 (pitch), and -64 degree to +70 degree (roll).

The 3DsMax software \cite{Ref33} was used to generate a generic 3D facial model as the initial model. The mean absolute error (MAE) of head pose angles between the ground-truth and the estimated ones and the corresponding standard deviation (STD) of the MAEs are computed to evaluate the estimation accuracy. To guarantee that the morphed 3D facial model is also symmetric with the facial symmetry plane, only the azimuth was morphed for the chin and the tip of nose, i.e., ${\Delta _{{\rm{1,}}\phi }}{\rm{ = 0}}$ and ${\Delta _{{\rm{2,}}\phi}}{\rm{ = 0}}$; while for the left and right canthus, both the azimuth and elevation were morphed with constraints ${\Delta_{{\rm{3,}}\varphi }}={\Delta _{{\rm{4,}}\varphi }}{\rm{ }}$ and ${\Delta _{{\rm{3,}}\phi }}{\rm{}}=-{\Delta _{{\rm{4,}}\phi }}{\rm{}}$.

\begin{figure}
\centering
\includegraphics[width=8.8cm]{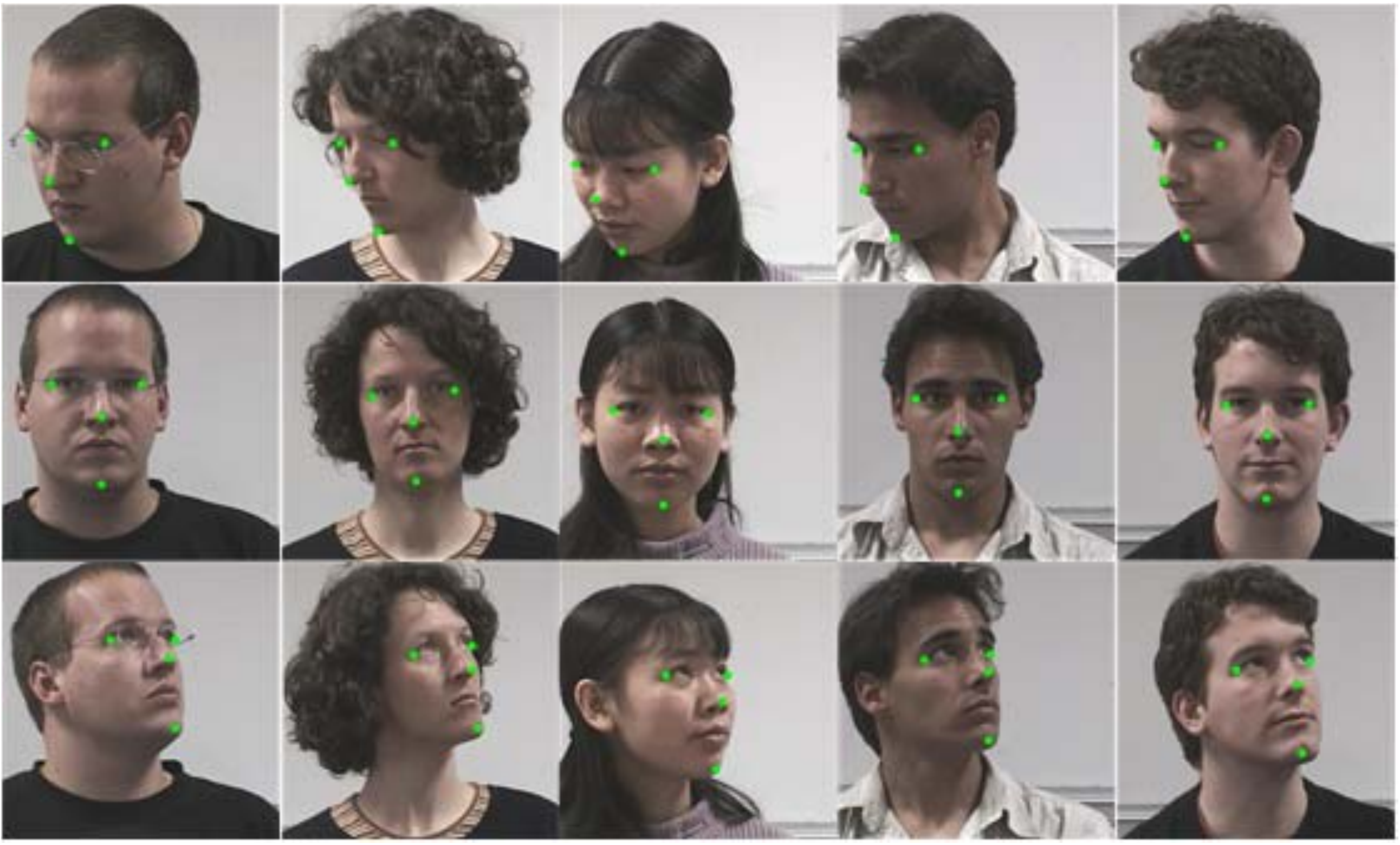}
\caption{Feature points used in the experiments, i.e. the chin, the
tip of nose, and the left and right canthus.} \label{fig3}
\end{figure}

\subsection{Head pose estimation with Manually Labelled 2D Feature Points}
We evaluated the proposed method with manually labelled 2D feature points. As shown in Fig. 3, the four feature points that were manually labelled in 2D face images and used in our method are the chin, the tip of nose on the facial symmetry plane, and the left and right canthus that are symmetrical with the facial symmetry plane based on the symmetry property of human faces \cite{Ref34}. These four feature points are employed by also considering that they are less sensitive to facial expressions.

\begin{table}
\centering \caption{ACCURACY COMPARISON OVER THE \emph{POINTING'04}
DATASET} \label{table1}
\includegraphics[width=12cm]{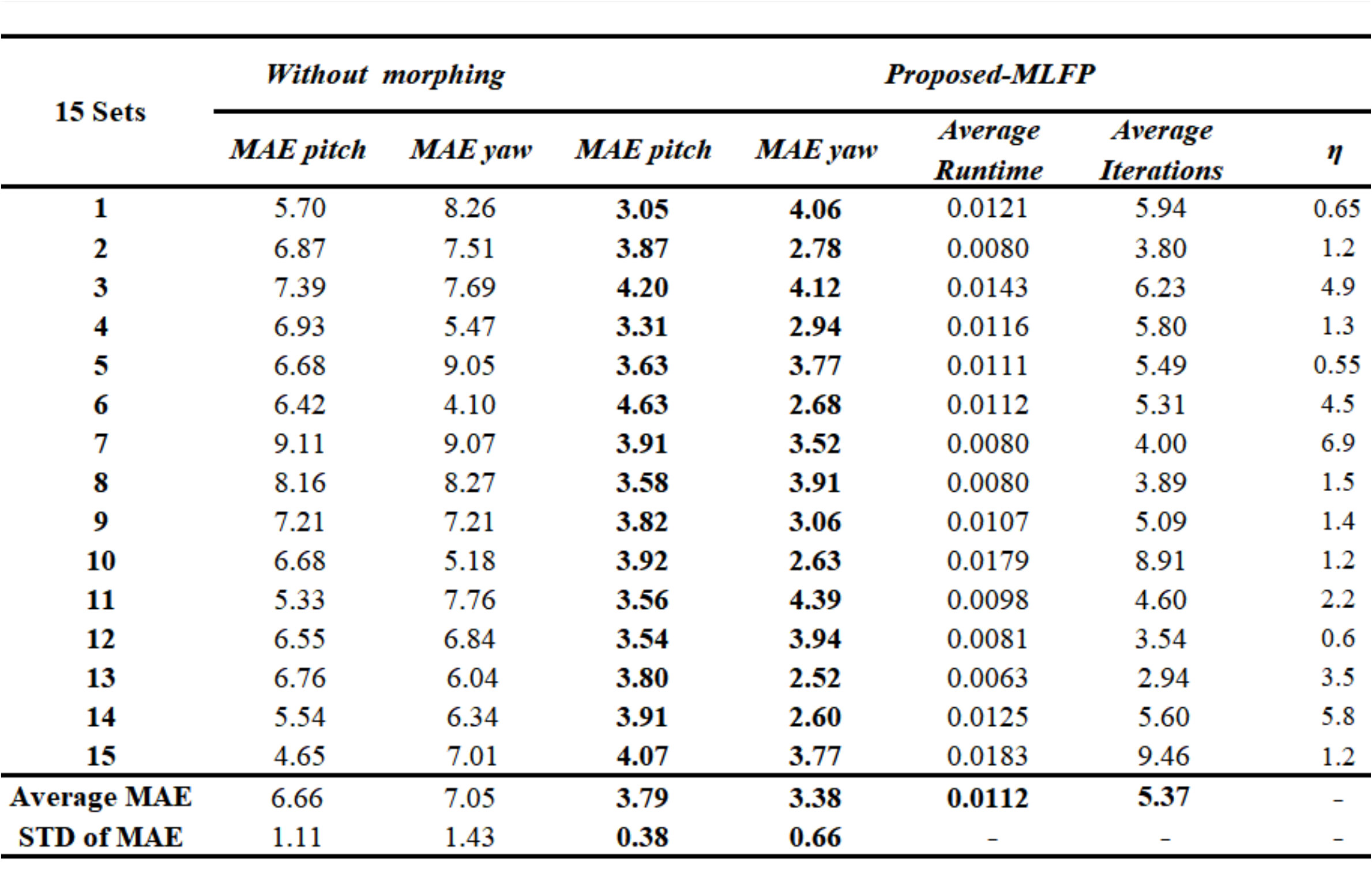}
\end{table}

The estimation errors are listed in Tables 1 and 2 for \emph{Pointing'04} and \emph{Biwi Kinect} datasets respectively. The plenty factor $\eta$ in (23), the average iteration numbers of images, and the average run times for images in each image sets of the dataset are also provided. From Table 1 (\emph{Pointing'04} dataset), we can observe that the average MAEs are 6.66 and 7.05 degree for the pitch and yaw angles when the 3D morphing operation is not conducted. For the proposed method (denoted as \textbf{\emph{Proposed-MLFP}}), the average MAEs are only 3.79 and 3.38 degree for the pitch and yaw angles respectively. From Table 2 (\emph{Biwi Kinect} dataset), we can observe that the average MAEs are 5.63, 3.72, and 2.09 degree for the pitch, yaw, and roll angles when the 3D morphing operation is not conducted. For \textbf{\emph{Proposed-MLFP}}, the corresponding average MAEs are only 5.04, 3.59, and 2.08 degree respectively. Besides, we can also observe that the STDs of MAEs of \textbf{\emph{Proposed-MLFP}} are also small, which means that the proposed method is more robust for different faces.

The variable $\eta$ is a penalty parameter to control the difference (or similarity) between the initial 3D facial model and deformed one during the iterative morphing such that the deformed one is still a facial model. The larger $\eta$ is, the more similar they are; that is, the initial 3D facial model will not be greatly deformed. Theoretically, the parameter $\eta$ should vary for different sets. The reason is that different sets consist of face images with different genders, races, ages, and facial expressions, the underlying 3D facial shapes of which change a lot from one set to another, i.e., the similarity between the initial 3D facial model and each ground-truth shape is different. Therefore, in the above experiments, the parameter has been fine tuned to nearly optimal under each set. From the perspective of practical applications, a fixed $\eta$ for all sets would be preferable. To this end, we set $\eta$ to 1.77 for all set, and the detailed results of each set by our method are listed in the following Table 3, where it can be observed that the average MAEs of pitch and yaw are 4.71 and 3.73, respectively, which are slightly larger than those by optimally tuning $\eta$ for each set separately, i.e., 3.79 and 3.38, demonstrating its practicality.

\begin{table}
\centering \caption{ACCURACY COMPARISON OVER THE \emph{BIWI KINECT}
DATASET} \label{table2}
\includegraphics[width=10cm]{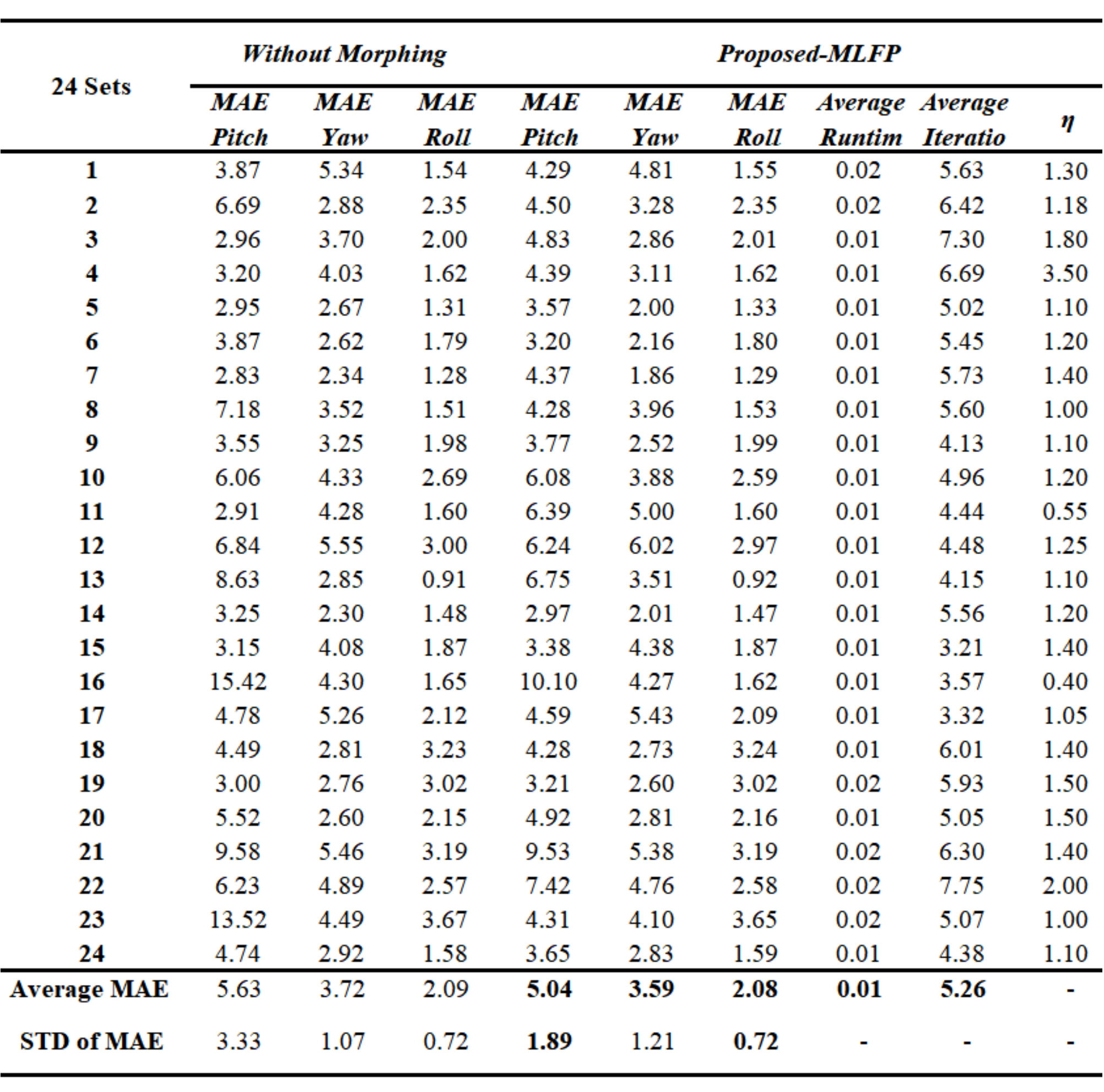}
\end{table}

For the proposed method, we only need four non-coplanar feature points which are not restricted to the chin, the tip of nose, and the left and the right canthus. When these feature points are invisible due to large pitch or yaw, other non-coplanar feature points could be used as an alternative. For example, we can choose the center of ears, the tip of nose, and the chin as feature points for larger pitch angles (-60, 60), and the chin, the tip of nose, and the left and the right canthus for larger yaw angles (-75, +75) (when the left or the right canthus are invisible, the leftmost or the rightmost point in the binocular connection line is used). We also conducted experiments to quantitatively verify the effectiveness of the proposed method for large pitch and yaw angles. As shown in Table 4, we can see that the proposed method still works very effectively, i.e., the average MAEs of both pitch and yaw are around 6.0.

\begin{table}
\centering \caption{EXPERIMENTAL RESULTS OF PROPOSED-MLFP WITH A
FIXED $\eta$ TO ALL SETS OF \emph{POINTING'04} DATASET ($\eta$=1.77).}
\label{table3}
\includegraphics[width=8cm]{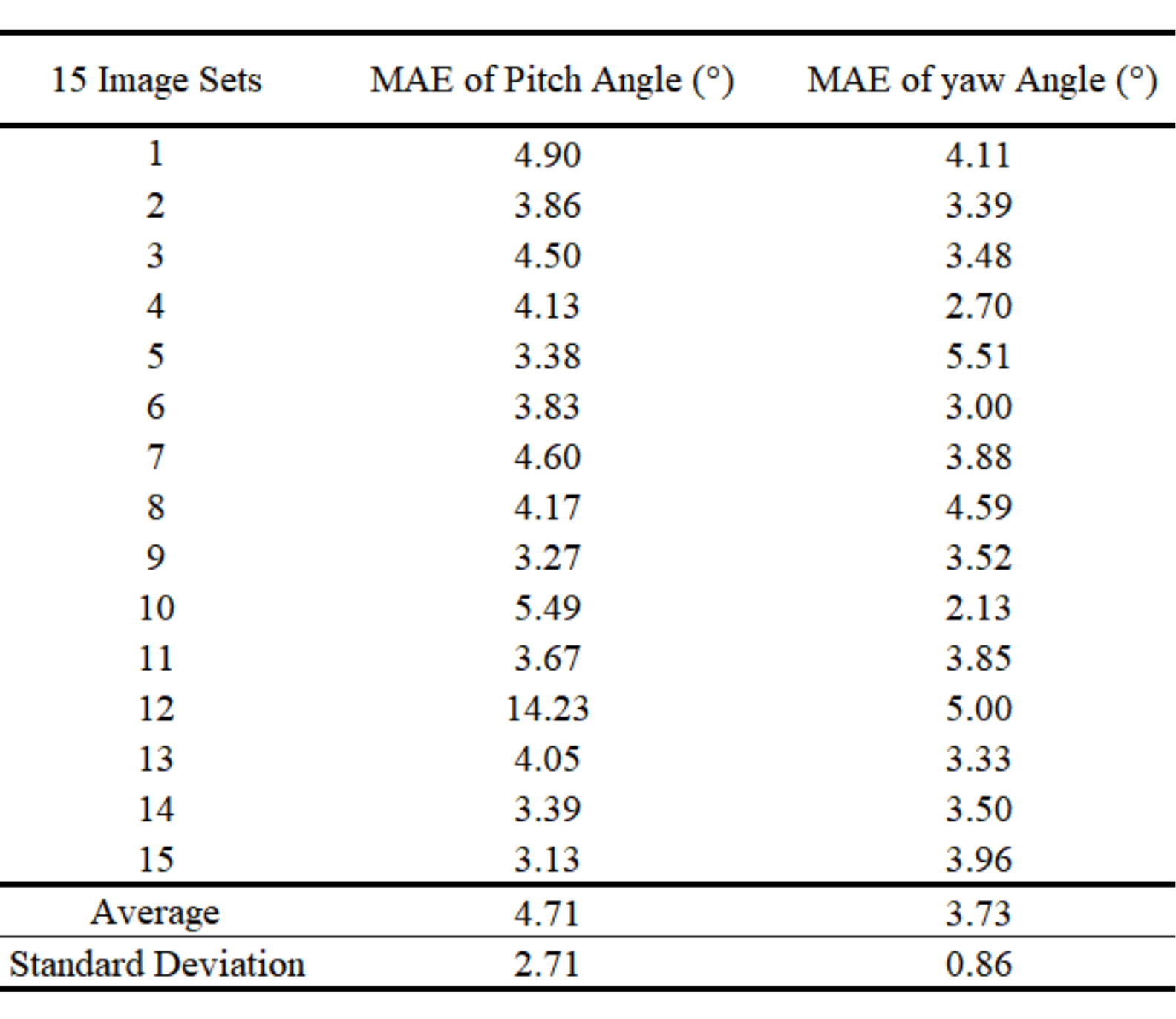}
\end{table}

\begin{table}
\centering \caption{EVALUATION OF THE PROPOSED METHOD FOR LARGE
PITCH AND YAW ANGLES. HERE, THE 8th SET IN \emph{POINTING'04} DATASET ARE
USED} \label{table4}
\includegraphics[width=12cm]{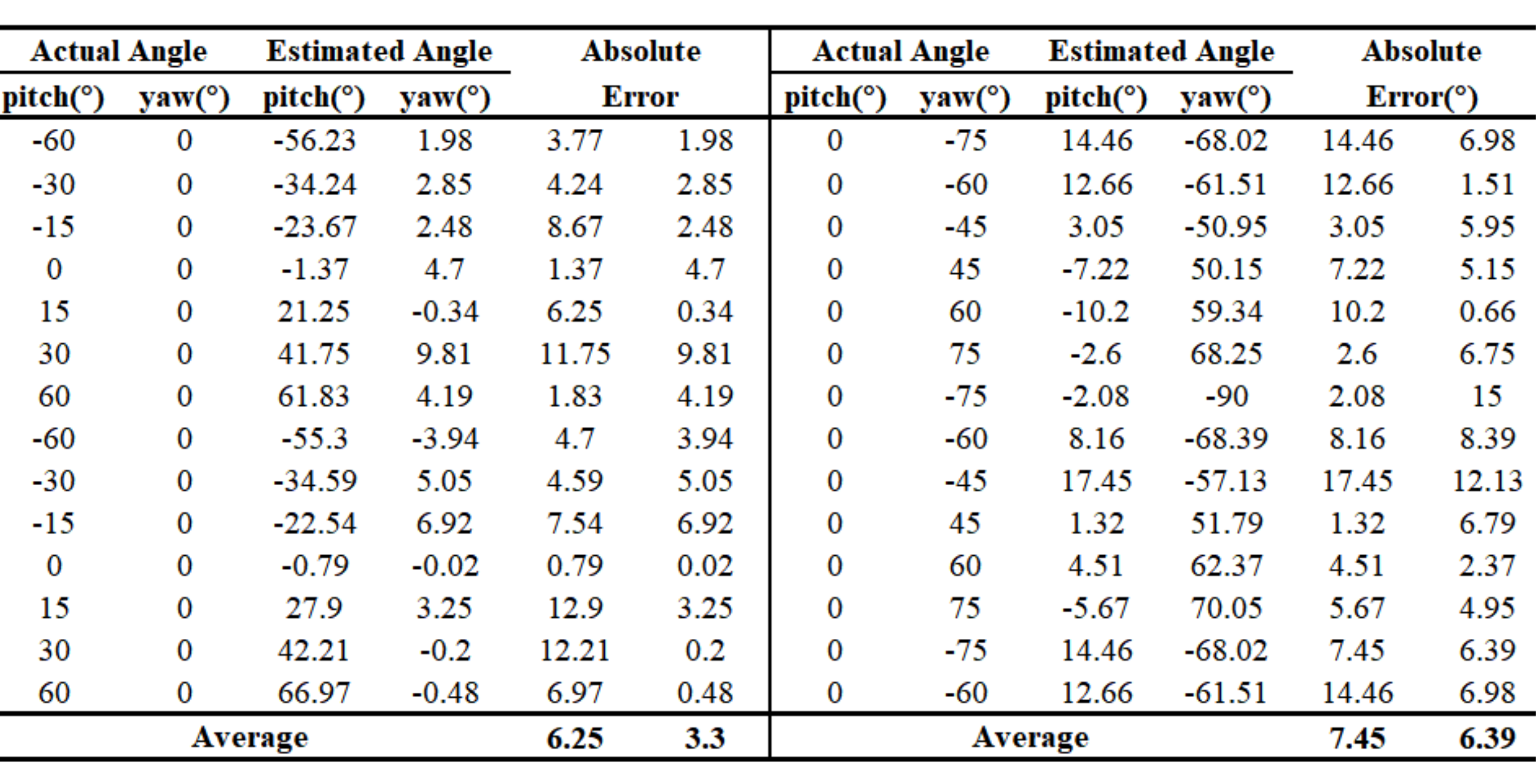}
\end{table}

\begin{table}
\centering \caption{ACCURACY COMPARISON BASED ON \emph{POINTING'04}
DATASET} \label{table5}
\includegraphics[width=12cm]{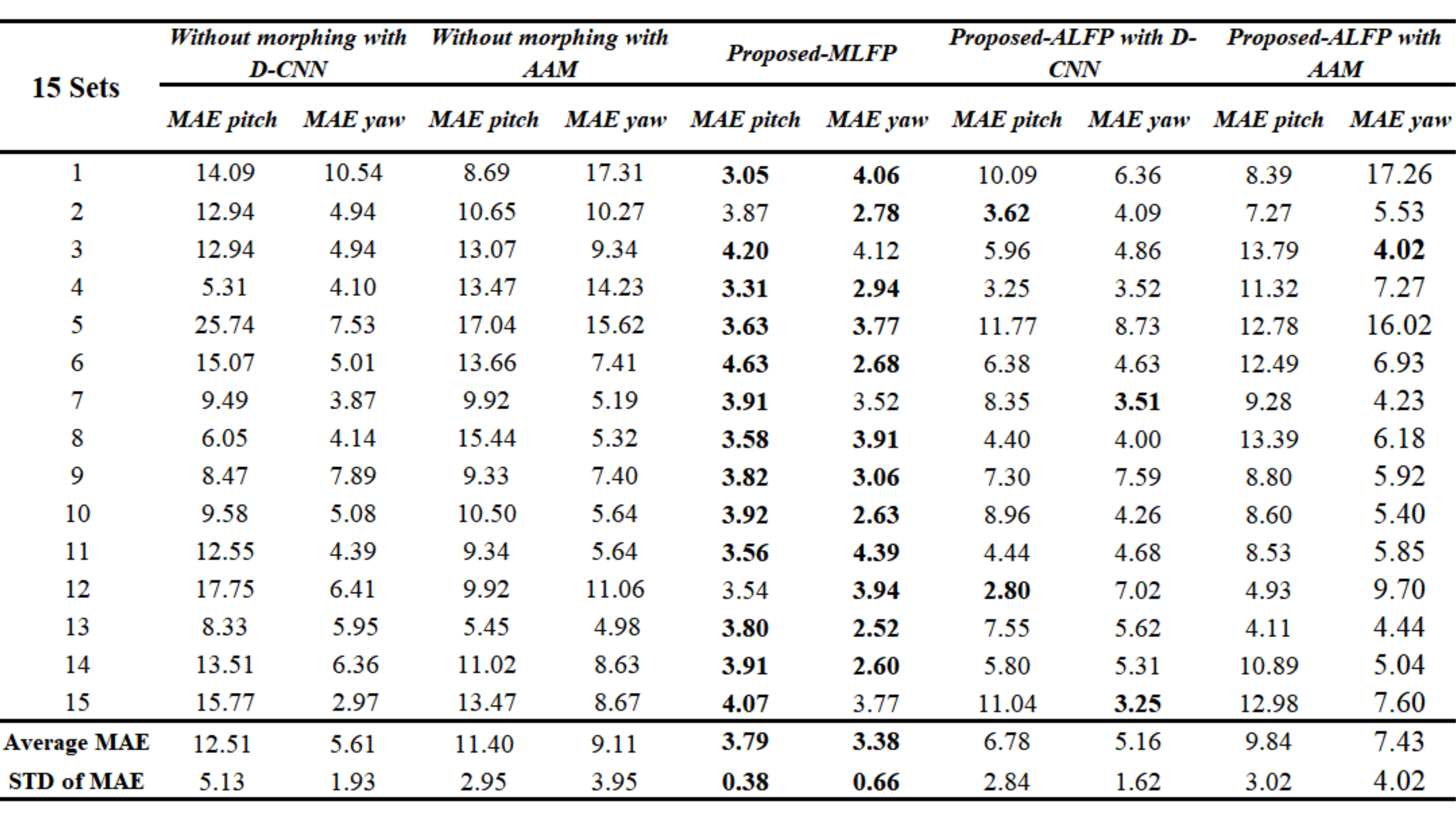}
\end{table}

\subsection{Head pose estimation with automatically labelled 2D feature points}
We also evaluated the proposed method with automatically labelled 2D feature points. The Deep Convolutional Network Cascade (D-CNN) \cite{Ref35} was adopted to detect feature points on 2D face images, in which five feature points, i.e., the two focal points of the eye, the tip of nose, and the two corners of the mouth, can be detected. Since only four feature points are needed in the proposed method, the two focal points of the eye, the tip of nose, and the midpoint of the two corners of the mouth are selected, as shown in the green points in Fig. 4.

Table 5 (\emph{Pointing'04} dataset) compares the estimation errors between the proposed method with automatically labelled 2D feature points (denoted as \textbf{\emph{Proposed-ALFP with D-CNN}}) and that with manually labelled feature points. We can see that the MAEs of \textbf{\emph{Proposed-ALFP with D-CNN}} are almost larger than those of \textbf{\emph{Proposed-MLFP}} for all the tested images. The reasons are that the midpoint of the two corners of the mouth may not be always located on the actual midpoint of mouth especially when the head pose and the facial expression is changed, as shown in the red points in Fig. 4.

Besides, we also used the AAM method \cite{Ref36} to extract 2D feature points. The corresponding estimation errors are also given in Table 5, from which we can observe that the accuracy of the \textbf{\emph{Proposed-ALFP with D-CNN}} is higher than that of the \textbf{\emph{Proposed-ALFP with AAM}} because D-CNN can extract feature points from 2D face images with higher accuracy than AAM.

\begin{figure}
\centering \label{fig4}
\includegraphics[width=7cm]{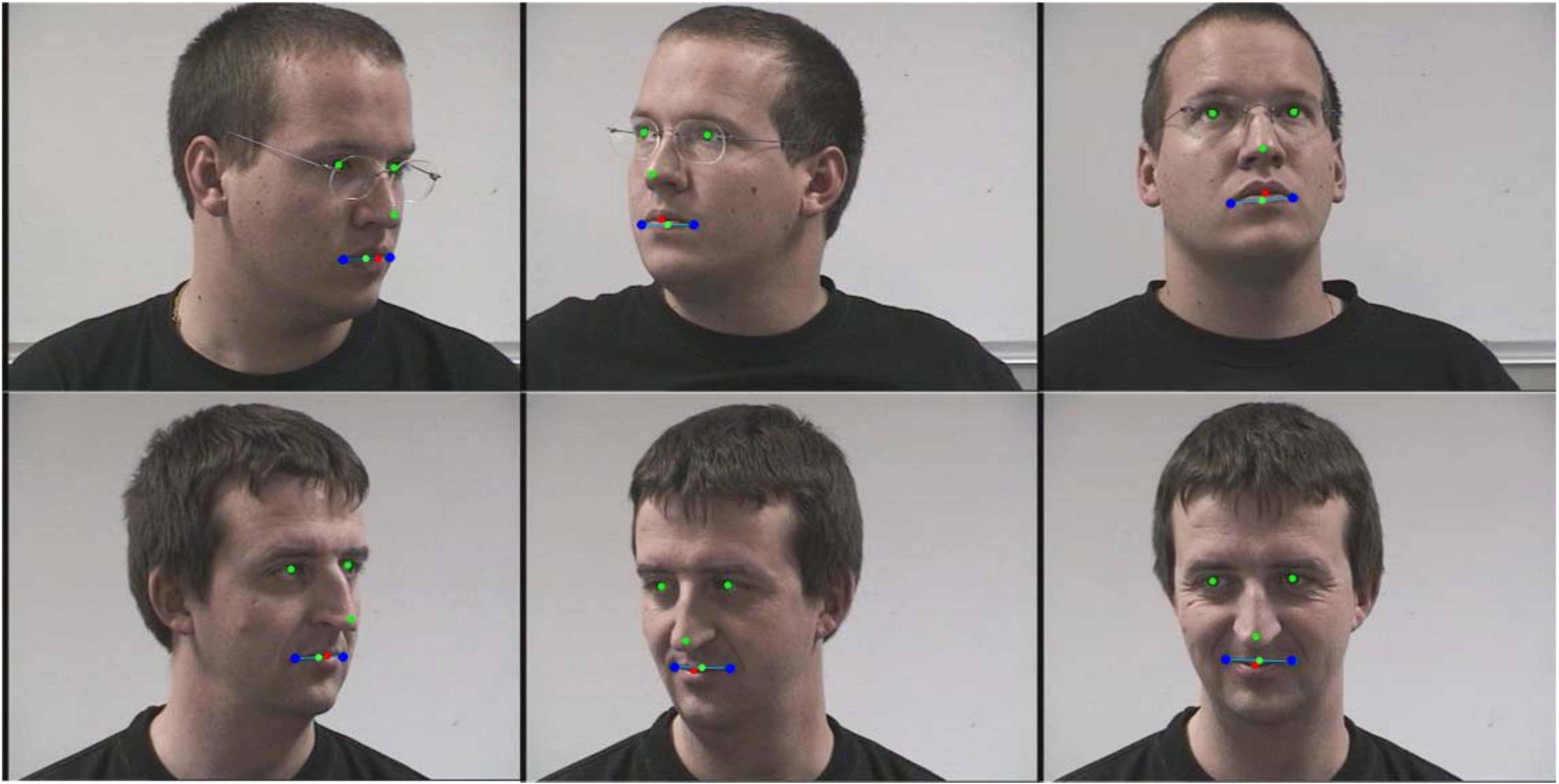}
\caption{The distribution of the midpoint of two mouth corners and
the actual midpoint (the red point) of mouth.}
\end{figure}

Moreover, Fig. 5 shows the convergence of estimation errors with respect to iterations. From this figure, we can observe that the estimation errors of the \textbf{\emph{Proposed-MLFP}} converges faster than that of the \textbf{\emph{Proposed-ALFP with AAM}}, while the converged estimation error of the \textbf{\emph{Proposed-ALFP with AAM}} is larger than that of the \textbf{\emph{Proposed-MLFP}}, which means that the performance of the proposed method depends on the accuracy of the 2D feature points.

\begin{figure}[htbp]
\centering \subfigure[]{
\begin{minipage}{7cm}
\centering
\includegraphics[width=7cm]{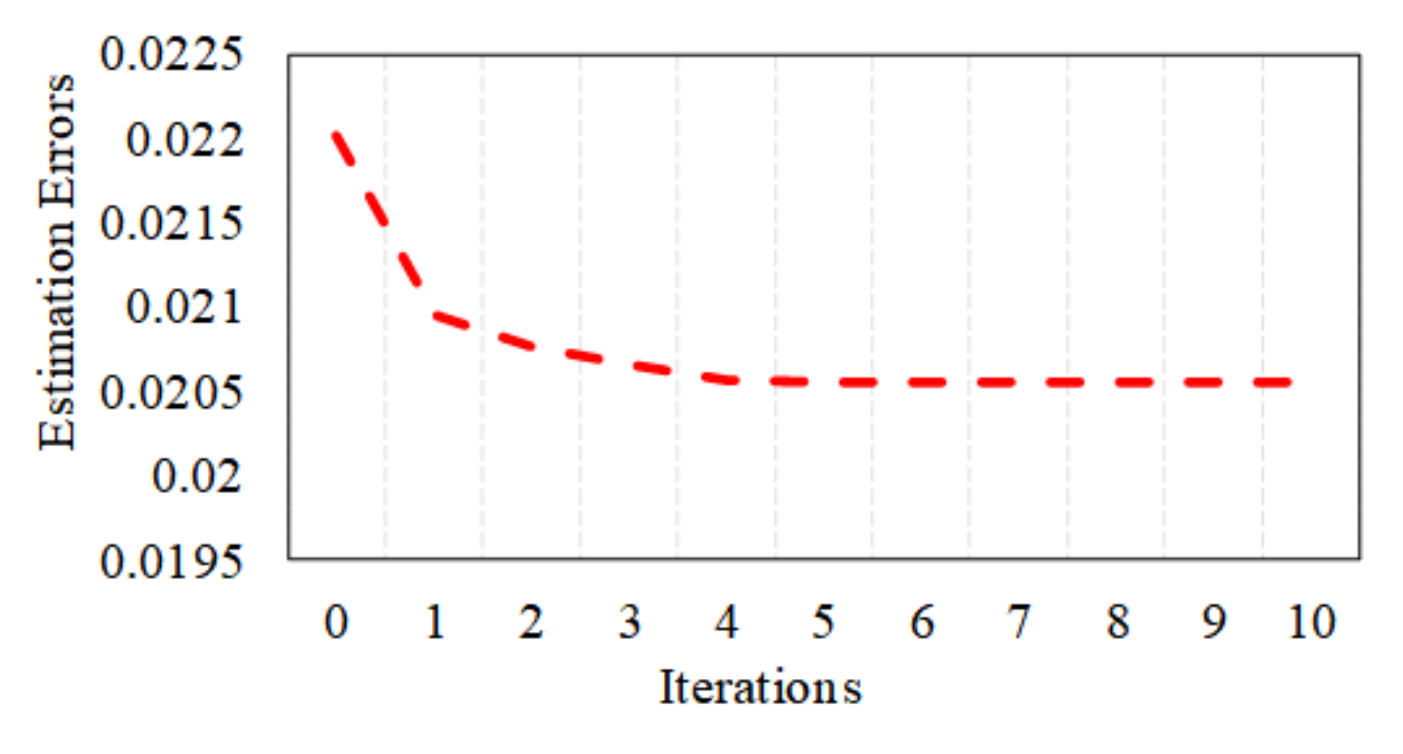}
\end{minipage}
}
\subfigure[]{
\begin{minipage}{7cm}
\centering
\includegraphics[width=7cm]{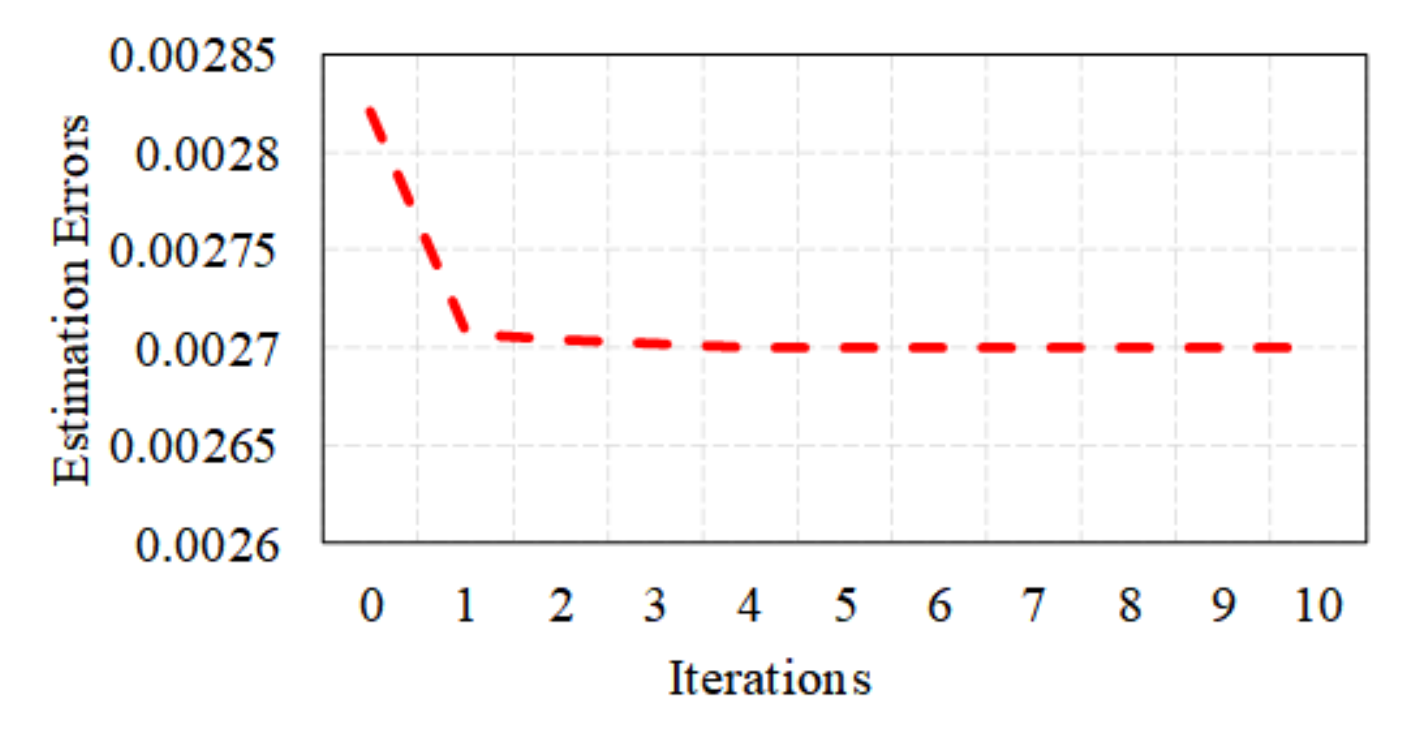}
\end{minipage}
} \caption{Estimation errors with respect to iterations,
(a)\textbf{\emph{Proposed-ALFP with AAM}},
(b)\textbf{\emph{Proposed-MLFP}}.} \label{fig:5}
\end{figure}

For the \emph{Biwi Kinect} dataset, the estimation errors of all the 24 sets are also compared in Table 6. From Table 6, we can observe that the average MAEs of pitch, yaw, and roll angles of \textbf{\emph{Proposed-ALFP with AAM}} are 6.70, 5.20, and 1.91, respectively. We can also observe that the STDs of MAEs of the \textbf{\emph{Proposed-ALFP with AAM}} are 3.55, 1.72, and 0.81 respectively. But, the average MAEs of pitch, yaw, and roll angles of \textbf{\emph{Proposed-ALFP with D-CNN}} are 5.94, 6.64, and 2.08, while the corresponding STDs of the MAEs are only 1.39, 1.21, and 0.73.

\begin{table}
\centering \caption{ESTIMATION ERRORS OF PITCH, YAW AND ROLL ANGLES
OF ALL THE 24 SETS IN \emph{BIWI KINECT} DATASET} \label{table6}
\includegraphics[width=12cm]{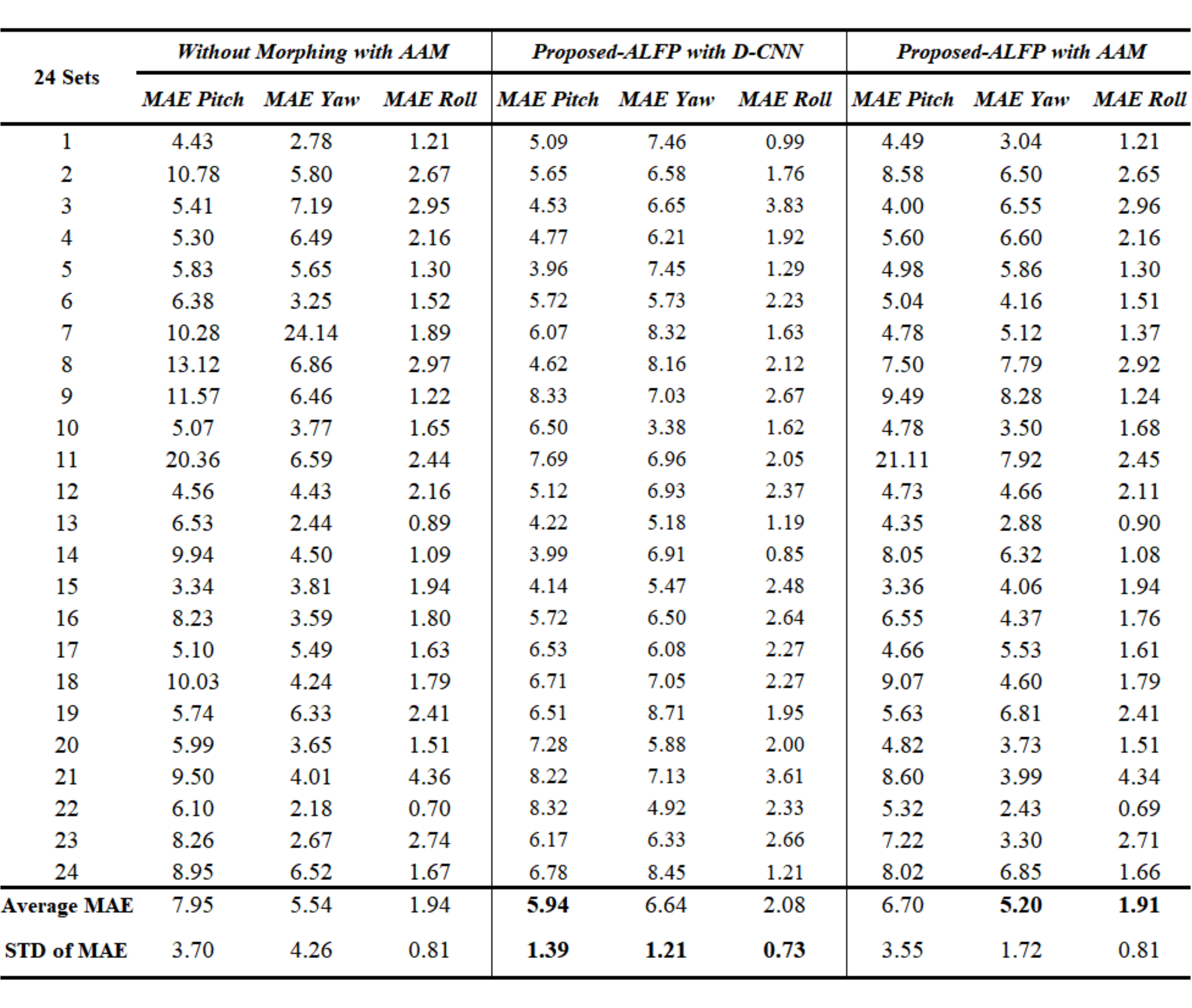}
\end{table}

\begin{table}
\centering \caption{COMPARISONS WITH RECENT METHODS IN
\emph{POINTING'04} DATASET} \label{table7}
\includegraphics[width=12cm]{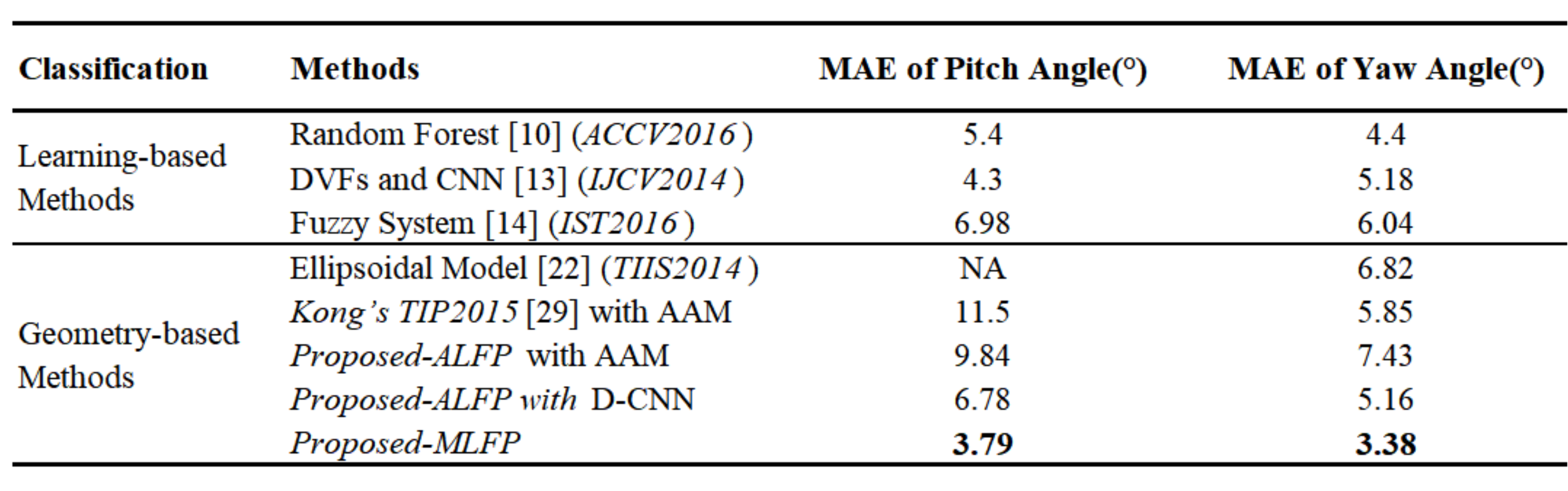}
\end{table}
\begin{table}
\centering \caption{COMPARISONS WITH RECENT METHODS IN \emph{BIWI
KINECT} DATASET} \label{table8}
\includegraphics[width=12cm]{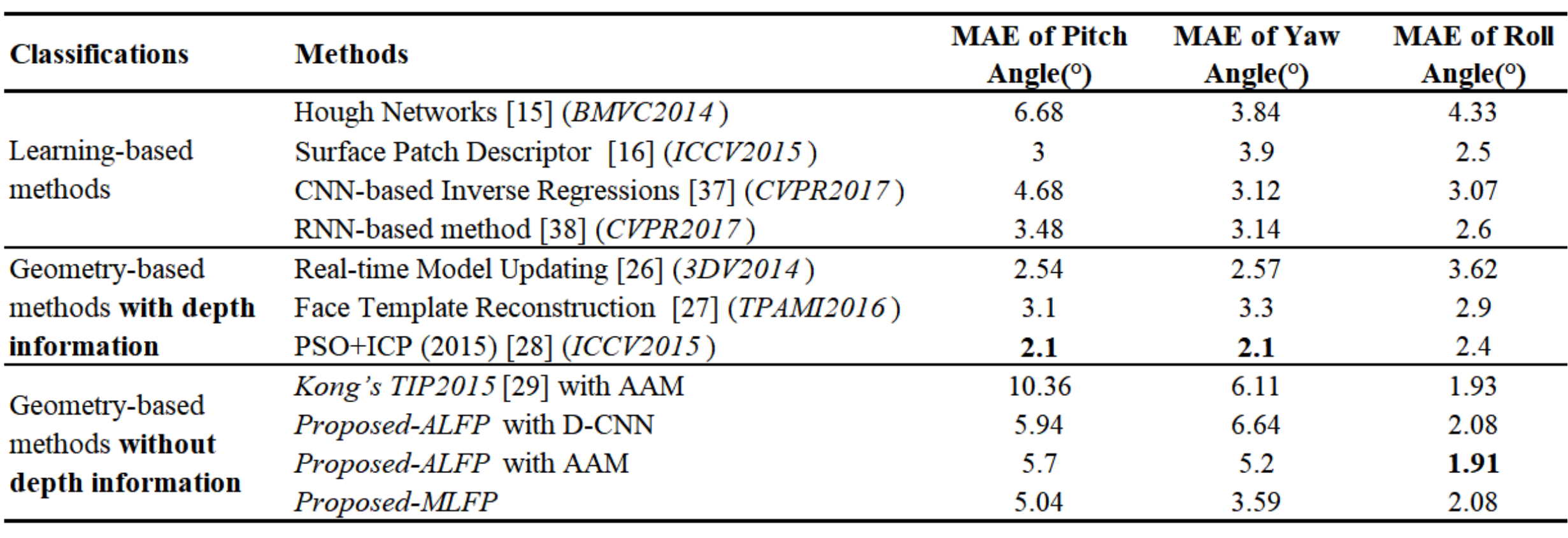}
\end{table}

Furthermore, in order to evaluate the performance of the proposed method comprehensively, we have compared the proposed method with some state-of-the-art methods, i.e., the improved Hough-voting with random forest \cite{Ref10}, the DVFs and CNN-based method in \cite{Ref13}, the fuzzy systems-based method in \cite{Ref14}, the ellipsoidal model-based geometry method in \cite{Ref22}, and the method in \cite{Ref29} in Table 7 for the \emph{Pointing'04} dataset, from which we can see that the MAE of the pitch and yaw angle of \textbf{\emph{Proposed-MLFP}} is the smallest, while the MAE of the pitch and yaw angle of \textbf{\emph{Proposed-ALFP with D-CNN}} is comparable to the start-of-the-art learning-based methods.

For the \emph{Biwi Kinect} dataset, we also compared the proposed method with some recent learning-based and geometry-based methods, including HNs-based method in \cite{Ref15}, surface patch descriptor-based method in \cite{Ref16}, real-time 3D facial model updating-based method in \cite{Ref26}, the face template reconstruction-based method in \cite{Ref27}, PSO and the ICP algorithm in \cite{Ref28}, CNN-based inverse regressions in \cite{Ref37} and Recurrent Neural Network (RNN)-based method \cite{Ref38}. The corresponding results are listed in Table 8, from which we can observe that the proposed method is comparable with those methods. In general, the performance of the proposed geometry-based method (without using additional depth information) is not as good as that of the learning-based methods. Interestingly, however, the accuracy of the roll angle of the proposed method is a little better than the learning-based methods. We believe the reason is that the range of the roll angle is very limited compared with the yaw and pitch angles. Therefore, the roll angle can be estimated more easily than the other two angles. For the learning-based methods, the train loss is usually set to the mean absolute error (MAE) of all the three directions (pitch, yaw, and roll), rather than a single direction. As a result, the accuracy of the roll angle may be compromised by the other two angles during the learning procedure.

\subsection{Complexity discussion}
When reporting the runtime in Tables 1, 2, and 6, the time of feature point detection have not been included. In practical applications, it would be more reasonable to take this time into consideration. Table 9 shows the run time of the CNN-based and AAM-based feature point detection methods. We can see that the 2D feature point detection phase that takes around 0.6 seconds is much more time-consuming than the proposed head pose estimation method (i.e., 0.01 seconds). Note that we have implemented the CNN-based feature point detection algorithms on a desktop with CPU@2.8GHz. We are confident that the whole pipeline has huge potential to run in real time (25 Hz+) by using GPU to accelerate the CNN-based feature point detection algorithm.

\begin{table}
\centering \caption{RUNTIME OF THE FEATURE POINT DETECTION METHODS
ON \emph{BIWI KINECT} DATASET} \label{table9}
\includegraphics[width=8cm]{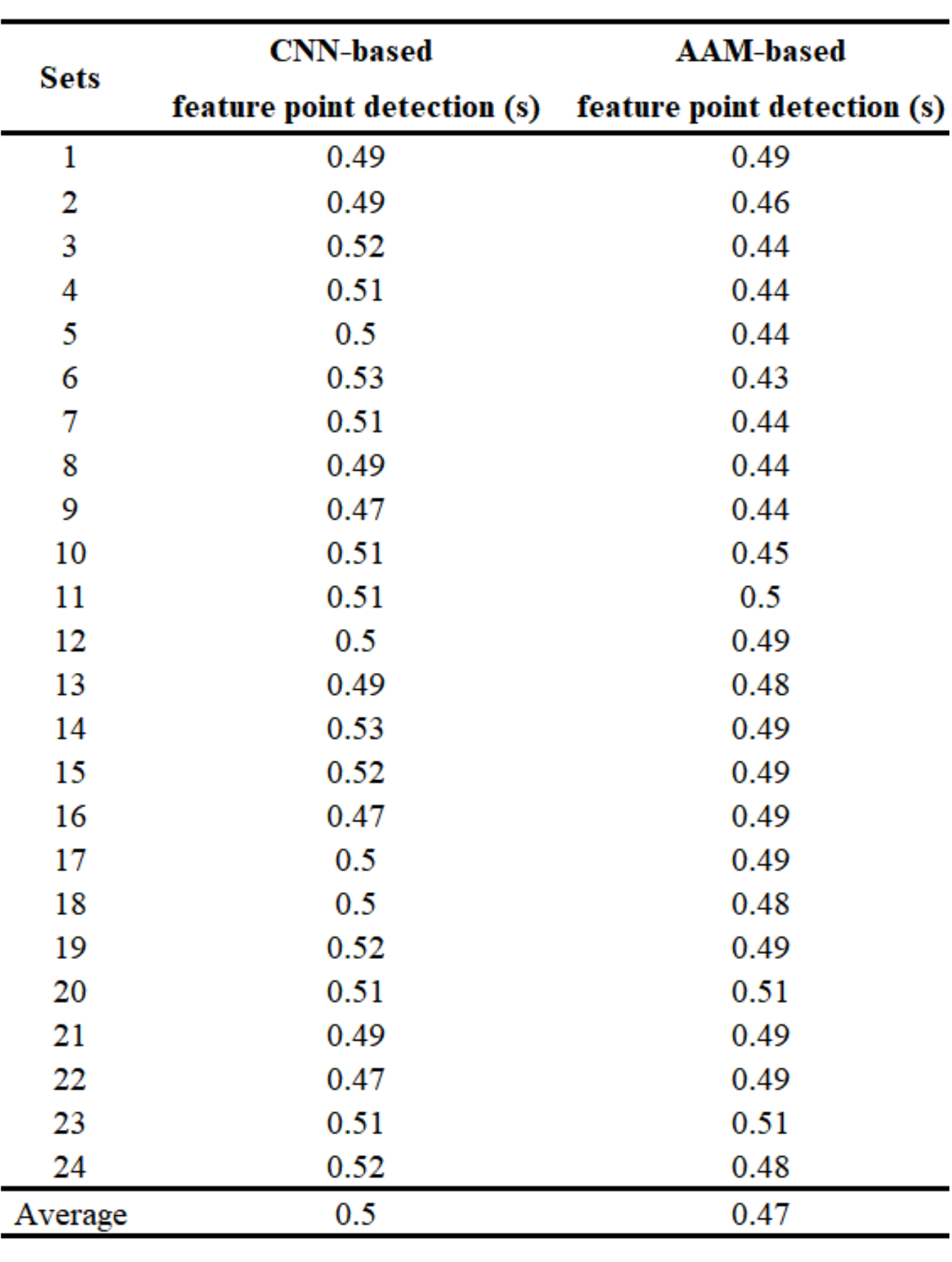}
\end{table}

\begin{table}
\centering \caption{COMPARISON WITH MOBILENET V3} \label{table10}
\includegraphics[width=12cm]{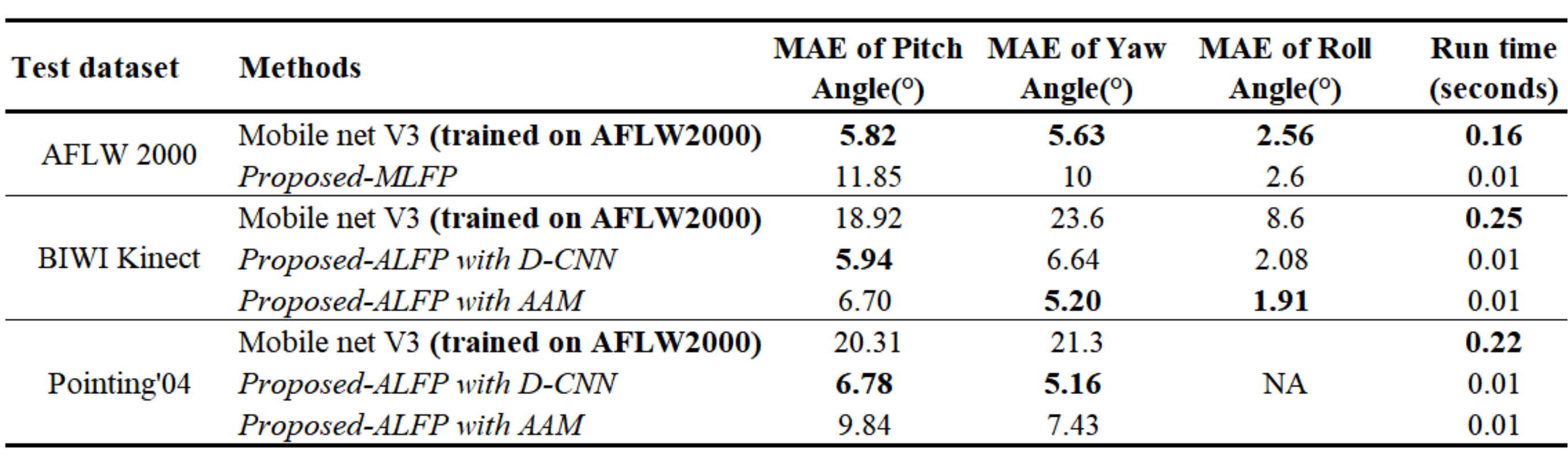}
\end{table}

\subsection{Additional experiments and analysis on the ALFW 2000 dataset}
Moreover, we also compared the proposed method with the most state-of-the-art lightweight CNN, i.e., MobileNet V3\cite{Ref39} by using the \emph{AFLW 2000} dataset\cite{Ref40}, and the results are given in Table 10. We can see that the accuracy (MAE) of MobileNet V3 is 5.82, 5.63, and 2.56 for the pitch, yaw, and roll angles, while the accuracy of the proposed method is 11.85, 10, and 2.6, respectively. In this case, the accuracy of the proposed method is much worse than the MobileNet V3-based approach. The reason is that the predefined 3D facial model in the proposed method cannot adapt to all the flexible 3D faces in the wild (as given in the \emph{AFLW 2000} dataset).

But we should note that the trained net model may not be suitable for the other datasets. In order to investigate the generalization ability of the MobileNet V3-based approach, we also tested it on the \emph{Pointing'04} dataset and \emph{BIWI Kinect} dataset by using the trained model based on the \emph{AFLW 2000} dataset. The results are also given in Table 10. We can see that, for the \emph{Pointing'04} dataset, the MAE of the pitch and yaw angles are 20.31 and 21.3, respectively, while for the \emph{BIWI Kinect} dataset, the MAE of the pitch, yaw, and roll angles are 18.92, 23.60, and 8.6, respectively, which are not good. However, the proposed method is still effective on these datasets.

Regard to the runtime, the average test time of a picture for MobileNet V3 is about 0.16 seconds, 0.25 seconds, and 0.22 seconds for the \emph{AFLW 2000}, the \emph{BIWI Kinect}, and the \emph{Pointing'04} datasets, respectively; while the runtime of the proposed method is only about 0.01 seconds.

\section{Conclusion}
We have presented a novel head pose estimation method based on the geometrical relationship between 4 non-coplanar feature points in 2D face image and those in a predefined 3D facial model. The coordinates of the 4 feature points are first converted to normalized coordinates so as to remove the influence of the scale and translation parameters in the 3D projection. Then the coordinates of feature points in the 3D facial model are transformed into the spherical coordinates and morphed to adapt various distributions of feature points of individual faces. The optimal morphing parameters with which the rotation matrix as well as head pose angles can be calculated are then found by the LM algorithm. Experimental results demonstrate that the accuracy of the roll angle estimated by the proposed method is the best when comparing with all the existing methods. For the pitch and yaw angles, the accuracy of the proposed method is the best when comparing with existing geometry-based method, and is comparable with existing methods that made used of depth information additionally.

\section*{Appendix 1}
\textbf{\emph{Proof of Eq. (19)}}:
According to (14), the relationships between $\textbf{\emph{m}}_0$ and $\textbf{\emph{M}}_0$, and between $\textbf{\emph{m}}_i$ and $\textbf{\emph{M}}_i$ can be respectively written as
\begin{equation}\label{E27}
\setcounter{equation}{27} \left\{ \begin{array}{l}
{\textbf{\emph{m}}_0} - {s{'}}{\textbf{\emph{t}}{'}} = {s{'}}{\textbf{\emph{R}}{'}}{\textbf{\emph{M}}_0}\left( i \right)\\
{\textbf{\emph{m}}_i} - {s{'}}{\textbf{\emph{t}}{'}} =
{s{'}}{\textbf{\emph{R}}{'}}{\textbf{\emph{M}}_i},\left( {ii}
\right)
\end{array} \right.
\end{equation}
where $\textbf{\textbf{t}}{'}=(t_1\ t_2)^T$. By subtracting Eq.(27)-(\emph{ii}) with Eq. (27)-(\emph{i}), we have
\begin{equation}\label{E28}
\setcounter{equation}{28} {\textbf{\emph{m}}_i} -
{\textbf{\emph{m}}_0} =
{s{'}}{\textbf{\emph{R}}{'}}({\textbf{\emph{M}}_i-\textbf{\emph{M}}_0}).
\end{equation}
Therefore, based on Eq. (15), $\textbf{\emph{m}}{'}_{i}$ can be rewritten as
\begin{equation}\label{E29}
\setcounter{equation}{29}
\textbf{\emph{m}}{'}_{i}=\frac{{s{'}}{\textbf{\emph{R}}{'}}({\textbf{\emph{M}}_i-\textbf{\emph{M}}_0})}{{\left\|
{\textbf{\emph{m}}_i-\textbf{\emph{m}}_0} \right\|_2}}.
\end{equation}
Furthermore, substitute $\textbf{\emph{M}}_i-\textbf{\emph{M}}_0$ by $\textbf{\emph{M}}{'}_{i}\cdot{{\left\| {\textbf{\emph{M}}_i-\textbf{\emph{M}}_0} \right\|_2}}$ in terms of Eq. (16), Eq. (29) can be represented as
\begin{equation}\label{E30}
\setcounter{equation}{30}
\textbf{\emph{m}}{'}_{i}=\frac{{s{'}}{\textbf{\emph{R}}{'}}\textbf{\emph{M}}{'}_{i}\cdot{{\left\|
{\textbf{\emph{M}}_i-\textbf{\emph{M}}_0} \right\|_2}}}{{\left\|
{\textbf{\emph{m}}_i-\textbf{\emph{m}}_0} \right\|_2}}.
\end{equation}
Accordingly, ${s{'}\cdot{{\left\|
{\textbf{\emph{M}}_i-\textbf{\emph{M}}_0} \right\|_2}}}$ must be
proved to be equal to ${{\left\|
{\textbf{\emph{m}}_i-\textbf{\emph{m}}_0} \right\|_2}}$.

Let $\emph{d}_0$ and $\emph{D}_0$ denote the ${{\left\|
{\textbf{\emph{m}}_i-\textbf{\emph{m}}_0} \right\|_2}}$ and
${s{'}\cdot{{\left\| {\textbf{\emph{M}}_i-\textbf{\emph{M}}_0}
\right\|_2}}}$ respectively, based on Eq. (28), we have
\begin{equation}\label{E31}
\begin{split}
\setcounter{equation}{31} \frac{{{d_0}}}{{{D_0}}}&=\frac{{{\left\|
{\textbf{\emph{m}}_i-\textbf{\emph{m}}_0}
\right\|_2}}}{{s{'}}\cdot{\left\|
{\textbf{\emph{M}}_i-\textbf{\emph{M}}_0} \right\|_2}}\\
&=\frac{{{\left\|
{s{'}}{\textbf{\emph{R}}{'}}({\textbf{\emph{M}}_i-\textbf{\emph{M}}_0})
\right\|_2}}}{{s{'}}\cdot{\left\|
{\textbf{\emph{M}}_i-\textbf{\emph{M}}_0} \right\|_2}}\\
&=\frac{{{\left\|
{\textbf{\emph{R}}{'}}({\textbf{\emph{M}}_i-\textbf{\emph{M}}_0})
\right\|_2}}}{{\left\| {\textbf{\emph{M}}_i-\textbf{\emph{M}}_0}
\right\|_2}},
\end{split}
\end{equation}
where ${{\left\|
{\textbf{\emph{R}}{'}}({\textbf{\emph{M}}_i-\textbf{\emph{M}}_0})
\right\|_2}}$ can be represented as
\begin{equation}\label{E32}
\begin{split}
\setcounter{equation}{32} &{{\left\|
{\textbf{\emph{R}}{'}}({\textbf{\emph{M}}_i-\textbf{\emph{M}}_0})
\right\|_2}}=\sqrt
{(\textbf{\emph{M}}_i-\textbf{\emph{M}}_0)^T{\textbf{\emph{R}}{'}}^T{\textbf{\emph{R}}{'}}(\textbf{\emph{M}}_i-\textbf{\emph{M}}_0)}\\
&=\sqrt{(\textbf{\emph{M}}_i-\textbf{\emph{M}}_0)^T{\left[\!{\begin{array}{*{20}{c}}
{{\textbf{\emph{r}}_1}}\
{{\textbf{\emph{r}}_2}}\end{array}}\!\right]}{\left[{\begin{array}{*{20}{c}}\!{\textbf{\emph{r}}_1^T}\
{\textbf{\emph{r}}_2^T}\!
\end{array}}\right]}(\textbf{\emph{M}}_i-\textbf{\emph{M}}_0)}\\
&=\sqrt
{(\textbf{\emph{M}}_i-\textbf{\emph{M}}_0)^T({\textbf{\emph{r}}_1}\textbf{\emph{r}}_1^T
+{\textbf{\emph{r}}_2}\textbf{\emph{r}}_2^T)(\textbf{\emph{M}}_i-\textbf{\emph{M}}_0)}.
\end{split}
\end{equation}
Let \textbf{\emph{C}} denotes
${\textbf{\emph{r}}_1}\textbf{\emph{r}}_1^T +
{\textbf{\emph{r}}_2}\textbf{\emph{r}}_2^T$, we have
\begin{equation}\label{E33}
\begin{split}
\setcounter{equation}{33}
\textbf{\emph{r}}_2^T\textbf{\emph{C}}&=\textbf{\emph{r}}_2^T({\textbf{\emph{r}}_1}\textbf{\emph{r}}_1^T
+ {\textbf{\emph{r}}_2}\textbf{\emph{r}}_2^T)\\
&=\textbf{\emph{r}}_2^T\textbf{\emph{r}}_1\textbf{\emph{r}}_1^T+\textbf{\emph{r}}_2^T\textbf{\emph{r}}_2\textbf{\emph{r}}_2^T.
\end{split}
\end{equation}
Because $\textbf{\emph{r}}_2^T\textbf{\emph{r}}_1=0$ and
$\textbf{\emph{r}}_2^T\textbf{\emph{r}}_2=1$, Eq. (33) can be
derived as
\begin{equation}\label{E34}
\setcounter{equation}{34}
\textbf{\emph{r}}_2^T\textbf{\emph{C}}=\textbf{\emph{r}}_2^T.
\end{equation}
That is to say \textbf{\emph{C}} must be an identity matrix.
Accordingly, Eq. (31) can be rewritten as
\begin{equation}\label{E35}
\setcounter{equation}{35} \frac{{{d_0}}}{{{D_0}}}=\frac{\sqrt
{(\textbf{\emph{M}}_i-\textbf{\emph{M}}_0)^T\textbf{\emph{C}}(\textbf{\emph{M}}_i-\textbf{\emph{M}}_0)}}{\sqrt
{(\textbf{\emph{M}}_i-\textbf{\emph{M}}_0)^T(\textbf{\emph{M}}_i-\textbf{\emph{M}}_0)}}=1.
\end{equation}
\textbf{\emph{Proof end\emph{.}}}

\section*{Appendix 2}
\textbf{\emph{Lemma}: A sphere can be uniquely determined by 4
non-coplanar points}.

\textbf{\emph{Proof}}: For a sphere centered at $(x_0\ y_0\ z_0)$ with radius of \emph{l}, the equation of any point $(x\ y\ z)$ on the sphere can be written as,
\begin{equation}\label{E36}
\setcounter{equation}{36} (x-x_0)^2+(y-y_0)+(z-z_0)=l^2.
\end{equation}
Therefore, in order to determine a sphere uniquely, the set of parameters ${\{x_0,y_0,z_0,l\}}$ must be calculated uniquely. When there are 4 non-coplanar points, i.e., $(x_i\ y_i\ z_i)$, $i\in\{1,2,3,4\}$, we should prove that there is a unique solution for the following equations:
\begin{equation}\label{E37}
\setcounter{equation}{37} \left\{ \begin{array}{l}
{\left( {{x_1} - {x_0}} \right)^2} + {\left( {{y_1} - {y_0}} \right)^2}+{\left( {{z_1} - {z_0}} \right)^2} = {l^2}\left( i \right)\\
{\left( {{x_2} - {x_0}} \right)^2} + {\left( {{y_2} - {y_0}} \right)^2}+{\left( {{z_2} - {z_0}} \right)^2} = {l^2}\left( {ii} \right)\\
{\left( {{x_3} - {x_0}} \right)^2} + {\left( {{y_3} - {y_0}} \right)^2}+{\left( {{z_3} - {z_0}} \right)^2} = {l^2}\left( {iii} \right)\\
{\left( {{x_4} - {x_0}} \right)^2} + {\left( {{y_4} - {y_0}}
\right)^2}+{\left( {{z_4} - {z_0}} \right)^2} = {l^2}\left( {iv}
\right)
\end{array} \right.
\end{equation}
By subtracting Eq. (37)-(\emph{i}) with Eq. (37)-(\emph{ii}), Eq. (37)-(\emph{iii}), Eq. (37)-(\emph{iv}), we have
\begin{equation}\label{E38}
\setcounter{equation}{38} \left\{\!\begin{array}{l}
{x_0}\!\left(\!{{x_2}\!-\!{x_1}}\!\right)\!+\!{y_0}\!\left(\!{{y_2}\!-\!{y_1}}\!\right)\!+\!{z_0}\!\left(\!{{z_2}\!-\!{z_1}}\!\right)\!=\!\frac{{\left( {x_2^2 + y_2^2 + z_2^2} \right) - \left( {x_1^2 + y_1^2 + z_1^2} \right)}}{2}\\
{x_0}\!\left(\!{{x_3}\!-\!{x_1}}\!\right)\!+\!{y_0}\!\left(\!{{y_3}\!-\!{y_1}}\!\right)\!+\!{z_0}\!\left(\!{{z_3}\!-\!{z_1}}\!\right)\!=\!\frac{{\left( {x_3^2 + y_3^2 + z_3^2} \right) - \left( {x_1^2 + y_1^2 + z_1^2} \right)}}{2}\\
{x_0}\!\left(\!{{x_4}\!-\!{x_1}}\!\right)\!+\!{y_0}\!\left(\!{{y_4}\!-\!
{y_1}}\!\right)\!+\!{z_0}\!\left(\!{{z_4}\!-\!{z_1}}\!\right)\!=\!
\frac{{\left( {x_4^2 + y_4^2 + z_4^2} \right) - \left( {x_1^2 +
y_1^2 + z_1^2} \right)}}{2}
\end{array} \right.
\end{equation}
which can be described as
$\textbf{\emph{P}}\cdot\textbf{\emph{S}}_\textbf{\emph{c}}=\textbf{\emph{b}}$:
\begin{equation}\label{E39}
\setcounter{equation}{39} \textbf{\emph{P}} = \left[
{\begin{array}{*{20}{c}}
{\left( {{x_2} - {x_1}} \right)}&{\left( {{y_2} - {y_1}} \right)}&{\left( {{z_2} - {z_1}} \right)}\\
{\left( {{x_3} - {x_1}} \right)}&{\left( {{y_3} - {y_1}} \right)}&{\left( {{z_3} - {z_1}} \right)}\\
{\left( {{x_4} - {x_1}} \right)}&{\left( {{y_4} - {y_1}} \right)}&{\left( {{z_4} - {z_1}} \right)}
\end{array}} \right]{,}
\end{equation}

\begin{equation}\label{E40}
\setcounter{equation}{40} \textbf{\emph{b}} = \left[
{\begin{array}{*{20}{c}}
{\frac{{\left( {x_2^2 + y_2^2 + z_2^2} \right) - \left( {x_1^2 + y_1^2 + z_1^2} \right)}}{2}}\\
{\frac{{\left( {x_3^2 + y_3^2 + z_3^2} \right) - \left( {x_1^2 + y_1^2 + z_1^2} \right)}}{2}}\\
{\frac{{\left( {x_4^2 + y_4^2 + z_4^2} \right) - \left( {x_1^2 +
y_1^2 + z_1^2} \right)}}{2}}
\end{array}} \right]{,}
\end{equation}
\begin{equation}\label{E41}
\setcounter{equation}{41} \textbf{\emph{S}}_c=(x_0\ y_0\ z_0)^T{,}
\end{equation}
where the matrix \textbf{\emph{P}} means is three non-coplanar vectors or lines such that any one of the three lines cannot be described by the other two. Thus, the rank of \textbf{\emph{P}} is 3 and $\textbf{\emph{det}}(\textbf{\emph{P}})\neq0$. Therefore, according to \emph{Cramer's Rule} [59], we can get a unique solution for $\textbf{\emph{P}}\cdot\textbf{\emph{S}}_\textbf{\emph{c}}=\textbf{\emph{b}}$

\textbf{\emph{Proof end}}.

\section*{Acknowledgment}
The authors would like to thank the editors and anonymous reviewers for their valuable comments.

\end{document}